\documentclass[10pt,twocolumn,letterpaper]{article}
\usepackage{cvpr}

\usepackage{amsfonts}
\usepackage{amsmath}
\usepackage{amssymb}
\usepackage{bbm}
\usepackage{enumitem}
\usepackage{graphicx}
\usepackage{xcolor}
\usepackage{multirow}
\usepackage{graphicx}
\usepackage{lipsum}
\usepackage[normalem]{ulem}
\useunder{\uline}{\ul}{}
\usepackage[section]{placeins}

\definecolor{cvprblue}{rgb}{0.21,0.49,0.74}
\usepackage[pagebackref,breaklinks,colorlinks,citecolor=cvprblue]{hyperref}

\def\paperID{911}
\def\confName{CVPR}
\def\confYear{2024}

\title{FastMAC: Stochastic Spectral Sampling of Correspondence Graph}
\author{
Yifei Zhang\textsuperscript{2,1},
Hao Zhao\textsuperscript{2}\footnotemark[2],
Hongyang Li\textsuperscript{3},
Siheng Chen\textsuperscript{4,3}\\
\textsuperscript{1}University of Chinese Academy of Sciences,
\textsuperscript{2}AIR, Tsinghua University, \\
\textsuperscript{3}Shanghai AI Laboratory,
\textsuperscript{4}Shanghai Jiao Tong University
\\
{\tt\small zhangyifei21a@mails.ucas.ac.cn, zhaohao@air.tsinghua.edu.cn}\\
}

\begin{document}
\twocolumn[{
\renewcommand\twocolumn[1][]{#1}
\maketitle
\begin{center}
    \captionsetup{type=figure}
    \includegraphics[width=0.80\textwidth]{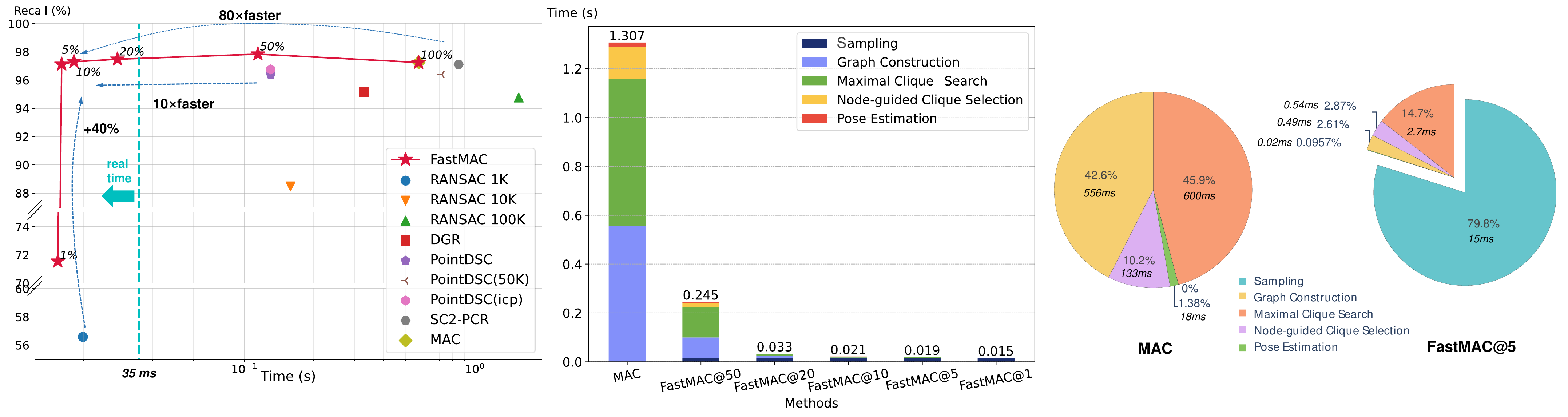}
    \captionof{figure}{\textbf{LEFT:} FastMAC can accelerate MAC \cite{zhang20233d} by 80 times, while preserving similarly high registration success rate (denoted by registration recall). This is achieved by sampling 5\% nodes on the correspondence graph, through a stochastic spectral formulation. Other sampling ratios are also shown, and FastMAC achieves real-time when the ratio is lower than 20\%. \textbf{MIDDLE:} Time profiling comparison between vanilla MAC and FastMAC with different sampling ratios. FastMAC significantly accelerates all stages of MAC. \textbf{RIGHT:} A detailed runtime breakdown for each component in MAC and FastMAC. Maximal clique search is no longer a bottleneck. }
    \label{fig:teaser}
\end{center}
}]

\footnotetext[2]{Corresponding author.}
\begin{abstract}
3D correspondence, i.e., a pair of 3D points, is a fundamental concept in computer vision. A set of 3D correspondences, when equipped with compatibility edges, forms a correspondence graph. This graph is a critical component in several state-of-the-art 3D point cloud registration approaches, e.g., the one based on maximal 
cliques (MAC). However, its properties have not been well understood.
So we present the first study that introduces graph signal processing into the domain of correspondence graph. We exploit the generalized degree signal on correspondence graph and pursue sampling strategies that preserve high-frequency components of this signal. To address time-consuming singular value decomposition in deterministic sampling, we resort to a stochastic approximate sampling strategy. As such, the core of our method is the stochastic spectral sampling of correspondence graph. As an application, we build a complete 3D registration algorithm termed as FastMAC, 
that reaches real-time speed while leading to little to none performance drop. 
Through extensive experiments, we validate that FastMAC works for both indoor and outdoor benchmarks. For example, FastMAC can accelerate MAC by 80 times while maintaining high registration success rate on KITTI. Codes are publicly available at \href{https://github.com/Forrest-110/FastMAC}{https://github.com/Forrest-110/FastMAC}.

\end{abstract}

\section{Introduction}


Correspondence is one of the most fundamental computer vision concepts, since it encodes important geometric relationships such as multi-view transformation (2D-2D \cite{liu2010sift} or 3D-3D \cite{teed2021raft}\cite{zhong20233d}\cite{zhong2022snake} correspondence) or single-view projection (2D-3D correspondence \cite{shotton2013scene}\cite{wu2022sc}). 3D correspondence, which is by definition a pair of matched 3D points, plays an important role in 3D registration \cite{zhang1994iterative} and downstream applications like SLAM \cite{cao2023representation}, 3D reconstruction \cite{curless1996volumetric}\cite{li2023lode} and 3D scene understanding \cite{chen2022pq}\cite{gao2023dqs3d}\cite{gao2023semi}\cite{tian2022vibus}. While the community has studied 3D correspondence for a long time, 3D correspondence graph is not yet well-understood.

In this graph, each vertex is a 3D correspondence, and the edge connectivity is usually defined according to the compatibility between two correspondences. For example, if a certain compatibility metric is higher than a threshold, an edge is active between two correspondences. This graph is indeed a critical component in state-of-the-art 3D registration methods like MAC \cite{zhang20233d}. MAC, as our baseline, searches for maximal cliques on this graph and estimates relative poses using compatible correspondences.

However, MAC can be quite slow with numerous input correspondences, with a single registration can cost more than one second, as shown in Figure.~\ref{fig:teaser}.
This makes it far from deployment for real-time applications such as SLAM. 
A natural idea 
is to downsample this graph for efficiency. We introduce the framework of graph signal processing \cite{chen2015discrete} to achieve this goal. Specifically, we use the generalized degree, \textit{i.e.}, the weighted edge sum, of each node in the graph as the graph signal and preserves its high frequency component with a graph filter, which is the Laplacian matrix in our work. 
We derive an optimal sampling strategy that best preserves the filtered high-frequency signal of interest.

\textbf{High-frequency.} The principle is that we need to sample nodes with fast change of generalized degree over the correspondence graph (short as \emph{high-frequency nodes} for brevity), since they are better suited for maximal clique search. The intuition is three-fold: (1) Every maximal clique must contain one high-frequency node; (2) Maximal cliques tend to contain three or more high-frequency nodes; (3) Cut-points, which are always high-frequency nodes, are contained in more than one maximal clique.

\textbf{Stochastic Sampling.} There exist well-established deterministic sampling methods to recover a certain signal of interest on a graph \cite{chen2015discrete}. But they involve iterative singular value decomposition operations which are time-consuming thus contradict the goal of accelerating MAC. As such, we derive a stochastic approximated sampling strategy that takes constant time w.r.t. sample number.

To summarize, we make these contributions: (1) For the first time, we introduce graph signal processing to 3D correspondence graph. (2) We propose a stochastic sampling strategy that selects high-frequency nodes on a correspondence graph. (3) We develop a full registration pipeline FastMAC upon the proposed sampling method, which is suited for maximal clique search. (4) FastMAC achieves trivial performance drop on indoor and outdoor datasets, while achieving 80 times acceleration to a real-time level.



\section{Related Works}

3D point cloud registration is important to many real-world problems, including pose estimation \cite{murali2022active}, SLAM \cite{pomerleau2013comparing}, and 3D reconstruction \cite{delmerico2018comparison}. However, globally optimal 3D point cloud registration is very challenging and existing works can be summarized into six primary categories:


\textbf{Maximum Consensus} (MC) is a widely used robust optimization objective in point cloud registration \cite{medioni2004emerging}. Compared with other robust optimization objectives such as Truncated-Least-Squares (TLS), MC has its advantage of being superior under certain circumstances\cite{chin2022maximum}. However, MC may produce an error-prone estimate depending on the input conditions \cite{carlone2022estimation}. Moreover, directly solving Maximum Consensus entails an NP-hard computational complexity, which has been confirmed by prior research \cite{chin2018robust}.


\textbf{Stochastic Techniques.} To address the complexity of solving the NP-hard Maximum Consensus problem, stochastic techniques have been proposed and Random Sample Consensus (RANSAC) is a well-known one among them. Numerous extensions and enhancements to RANSAC have been devised, aiming to improve its efficiency \cite{tordoff2002guided}\cite{chum2005matching}, accuracy \cite{torr2002bayesian}, and robustness \cite{konouchine2005amlesac}\cite{sun2021ransic}. However, it is still essential to recognize that the convergence speed of RANSAC exhibits an exponential relationship with the rate of outliers in the dataset \cite{bustos2017guaranteed}.


\textbf{Branch-and-Bound.} The Branch-and-Bound (BnB) \cite{bazin2013globally}\cite{olsson2008branch}\cite{zheng2011deterministically} algorithm stands as a fundamental technique in optimization and search problems, for registration. 
It can explore and assess all solution possibilities systematically and intelligently remove less promising ones, ensuring an optimal solution.
Still, it is important to note that BnB exhibits an exponential complexity concerning the problem size and the presence of outliers within the dataset \cite{bustos2017guaranteed}. 


\textbf{Mixed Integer Program.} BnB has been extended with Mixed-Integer Programming (MIP) \cite{li2009consensus} to speed up computation. But MIP itself also demonstrates that the incorporation of Linear-Matrix-Inequality Constraints significantly expedites computational processes \cite{speciale2017consensus}. Several avenues of exploration has been done including TEASER++ \cite{yang2020teaser}, Fast-Global-Registration (FGR) \cite{zhou2016fast} and other works \cite{yang2020graduated}. Nevertheless, it is noteworthy that the computational time still exhibits sensitivity to both the outlier rate and problem size, underscoring the need for further improvements. 




\textbf{Simultaneous Pose and Correspondence} (SPC) methods represent another prominent paradigm within the field of point cloud registration, with the pioneering work of the Iterative Closest Point (ICP) algorithm as a cornerstone \cite{besl1992method}. Over time, several robust extensions of the ICP method have been introduced \cite{granger2002multi}\cite{kaneko2003robust}\cite{chetverikov2005robust}\cite{maier2011convergent}\cite{segal2009generalized}. SPC methods are often lauded for their swiftness and precision, yet they do exhibit a notable vulnerability to local minima, a limitation that has been acknowledged \cite{pomerleau2013comparing}. Though global SPC methods like Go-ICP \cite{yang2015go} have been proposed, it is important to note that many global methods in the SPC paradigm still rely on Branch-and-Bound (BnB) techniques \cite{hartley2009global}\cite{breuel2003implementation}.



\textbf{Consistency Graph-based Methods.} Recently, several approaches based on consistency graph are proposed for point cloud registration, emphasizing the encoding of consistency among pairs of correspondences through the utilization of invariants \cite{enqvist2009optimal}\cite{mangelson2018pairwise}\cite{leordeanu2005spectral}\cite{yang2021sac} structured within a graph framework. Researchers have sought to enhance the efficiency of the search for these maximum cliques through the introduction of more efficient search algorithms \cite{parra2019practical} and various relaxations to the maximum clique problem \cite{lusk2021clipper}\cite{shi2021robin}. Previous research \cite{quan2020compatibility} has explored sampling of correspondences, but we address the problem from the perspective of graph signal processing for the first time.

\begin{figure*}[]
\centering
\includegraphics[width=1.00\textwidth]{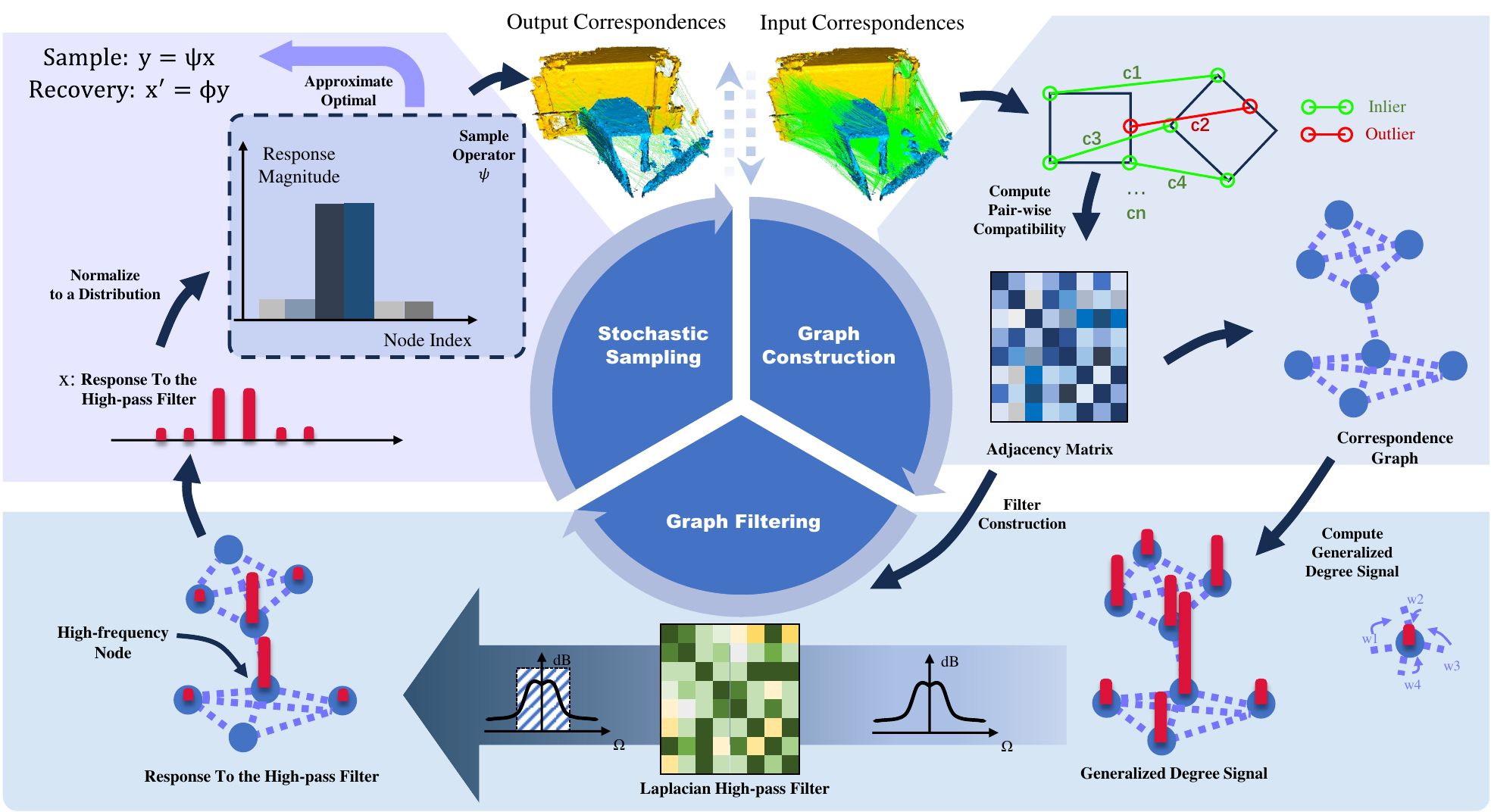}
\caption{
\textbf{Pipeline of FastMAC.}  In the \textbf{top-right} panel, we show the the procedure of constructing a correspondence graph from input correspondences. The graph is mathematically represented by an adjacency matrix. High values in this matrix mean high compatibility between two correspondences. In the \textbf{bottom} panel, we define generalized degree signal on the graph as the aggregation of compatibility scores on edges connecting with a node. We pass the signal through a Laplacian high-pass graph filter (constructed from the adjacency matrix) to get its high-frequency component. As mentioned in the text, after filtering, nodes with high response are named as \emph{high-frequency nodes}. In the \textbf{top-left} panel, we derive a stochastic sampling strategy in which the sampling probability of a node is proportional to the response magnitude. This sampling strategy is a fast approximation of the optimal (but slow) deterministic sampling strategy that recovers a signal of interest. \textbf{Lastly} but not shown in this figure, we use the MAC registration algorithm on output correspondences.
}
\label{fig:main-fig}
\end{figure*}

\section{Methods}\label{methods}

Our goal is fast and accurate 3D point cloud registration and our method is based upon the recently published maximal clique (MAC) method \cite{zhang20233d}. It has four steps as shown in the time profiling of Fig.~\ref{fig:teaser} and Table.~\ref{tab:profile}. (1) \textbf{Graph Construction}, which builds a correspondence graph on the input correspondences. We inherit this step as shown in the top-right panel of Fig.~\ref{fig:main-fig}. (2) \textbf{Maximal Clique Search}, which finds all maximal cliques in the graph as the name implies. (3) \textbf{Node-guided Clique Selection}, which reduces the number of maxmimal clique candidates and finally (4) \textbf{Pose Estimation}, which evaluates pose hypothesis generated in each clique and chooses the best one as the output pose. 

The key intuition behind MAC
is to loosen the previous maximum clique constraint \cite{lin2022k},
and use more maximal clique candidates to
generate potentially accurate pose hypotheses. However it is very slow when there are many input correspondences and \textbf{Maximal Clique Search} is the biggest bottleneck due to its exponential complexity. Hence, we aim to design a sampling module that reduces graph size without sacrificing the maximal clique registration performance. Our sampling module is shown in the bottom and top-left panels of Fig.~\ref{fig:main-fig}. It is inserted into the first and second step of MAC, which means the output correspondences are input into the remaining three steps of MAC. That is the difference between MAC and FastMAC.

How to achieve this graph down sampling? There are widely used modules like random sampling and farthest point sampling \cite{qi2017pointnet}. But as shown later in Fig.~\ref{fig:down1}, Fig.~\ref{fig:down2} and Fig.~\ref{fig:down3}, they perform poor for MAC acceleration. So we resort to the graph signal processing theory \cite{sandryhaila2014big}\cite{sandryhaila2014discrete}\cite{chen2015discrete}\cite{chen2017fast}. Due to page limit, its basics are presented in Appendix A.1.

First, as shown in the top-right panel of Fig.~\ref{fig:main-fig}, we construct a correspondence graph $G_{\rm corr}$ and the adjacency matrix $W_{\rm SOG}$, in which SOG means second order graph, following MAC \cite{zhang20233d}. Due to page limit, the details are presented in Appendix A.4.
It is noteworthy that the value in $W_{\rm SOG}$ means compatibility between two correspondences. 


\textbf{Generalized Degree Signal.} In order to exploit the graph signal processing theory, we need to define a signal on the correspondence graph. The normal degree signal for a node is the number of edges it connects. In a weighted graph, the generalized degree signal for a node is defined as the sum of edge weight it connects. In the following sections, we will explain why this signal can help us construct a graph filter that is suited for maximal clique search.

\subsection{Graph Filtering: Key Insight}
\begin{figure}[t]
\centering
\includegraphics[width=0.60\linewidth]{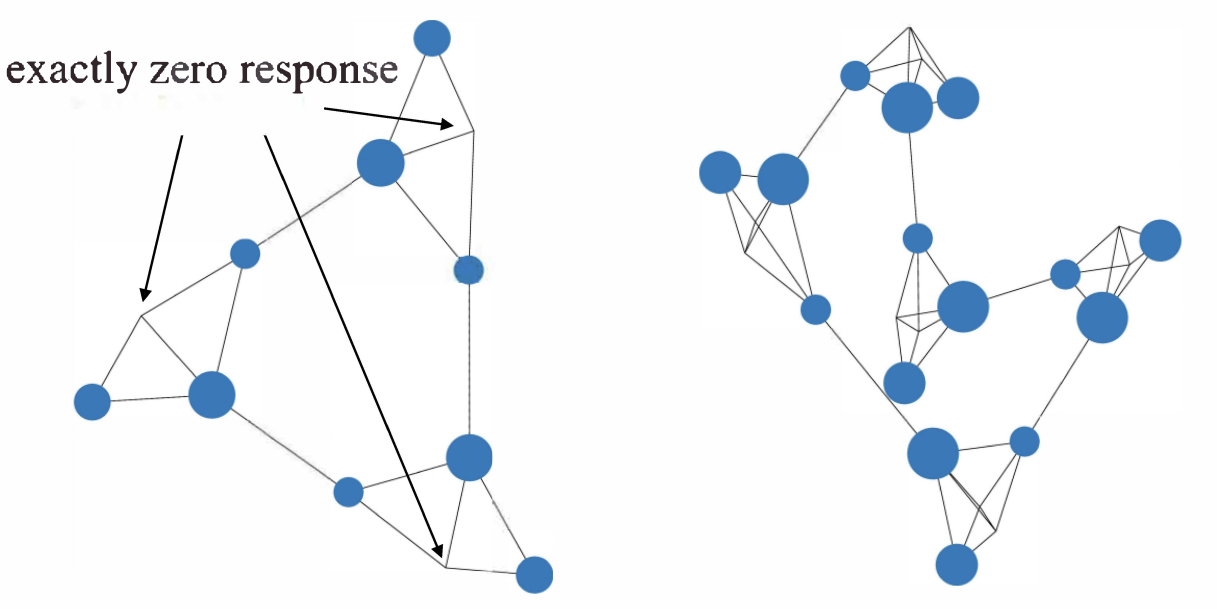}
\caption{
Response of the degree signal to a high-pass filter on connected-caveman graphs. Node size represents the response magnitude. This shows why high-pass filter is suited for MAC.
}
\label{fig:cave}
\end{figure}

After constructing a correspondence graph, we go into the second part of Fig.~\ref{fig:main-fig}. Our objective is to extract the high-frequency component of the generalized degree signal (degree later for brevity), allowing us to sample the nodes in the graph where the degree undergoes rapid changes. This section will highlight the reasons for doing so. We start by analysing the frequency of the degree distribution, and in particular its relation to cliques, as shown in Fig.~\ref{fig:cave}.

In order to explore the relationship between the degree signal frequency and cliques, we investigated the response of the degree signal to a high-pass filter, which is implemented with Laplacian matrix as mentioned above, on crafted (not generated from realistic data) connected caveman graphs\cite{watts1999networks}. 
Note that our focus is on the degree frequency, a local feature determined solely by the neighboring nodes. By prioritizing this local feature, we can simplify other types of graphs into the connected caveman graph, which will be explained later.

As illustrated in Fig.~\ref{fig:cave}, nodes with high response exhibit certain properties. If we consider each clique as a community, then:
1.In every community, there must exist a node to generate strong response. 
2.There are a sufficient number of nodes within a community that can elicit a strong response.
3.Nodes with significant response lie on the periphery of each community and are susceptible to consituting cut points. They have links not only with nodes within their respective community but also with nodes in other ones.

These properties prompt the idea of sampling the high-frequency nodes.
Maximal Clique registration process involves searching all maximal cliques in the correspondence graph, generating hypothesis for each maximal clique and selecting the best one. Suppose the output samples consist of high-frequency nodes, then: 1. since such nodes must exist in every community, they can cover nearly every maximal clique. 2. A sufficient number of samples in each clique guarantees the ability to generate a hypothesis. 3. Considering link between nodes represents compatibility, the selected correspondences are not only compatible with the correspondences within their own clique, but also with some others, indicating that these correspondences are more reliable and thus generating better hypotheses.

Now we explain why the features seen in the connected caveman graph can apply to other graphs. In a typical graph, cliques are either mutually connected or not. Mutually connected cliques maintain similar local properties of a connected caveman graph, whereas isolated cliques exhibit distinct features. Nevertheless, isolated cliques are rare and often negligible in the scenario of graph-based registration.

\subsection{Graph Filtering: Formulation}
\textbf{High-pass.} Guided by the aforementioned insight, we propose to selectively sample the high-frequency nodes by first formulating the graph filter. There are three typical graph filters: high-pass, low-pass and all-pass. A simple design of high-pass filter is a Haar-like high-pass
graph filter: \begin{equation}
\mathcal{H}=I-\mathcal{A}\\
=V\begin{bmatrix} 
    1-\lambda_1 & 0 & \dots & 0 \\
    0 & 1-\lambda_2 & \dots & 0 \\
    \vdots & &\ddots & \\
    0 & 0 &  \dots    & 1-\lambda_{N} 
    \end{bmatrix}V^{-1}    
\end{equation}
where \(\mathcal{A}\) is a normalized graph shift as defined in Appendix A.1.,
\(V\) and \(\lambda_i\) are the corresponding eigenvectors and eigenvalues. Note that \(\lambda_{\rm max}=1\) and if we order \(\lambda_i\) in a descending order, we have $1-\lambda_i\leq 1-\lambda_{i+1}$, indicating low frequency response attenuates and high
frequency response amplifies. A detailed interpretation is given in \cite{chen2017fast}. 

\textbf{Low-pass.} The opposite of this is a  the Haar-like low-pass graph filter, that is
\begin{equation}
    \begin{aligned}
        \mathcal{H}&=I+\frac{1}{|\lambda_{\rm max}|}\mathcal{A}\\
&=V\begin{bmatrix} 
    1+\frac{\lambda_1}{|\lambda_{\rm max}|} & 0 & \dots & 0 \\
    0 & 1+\frac{\lambda_2}{|\lambda_{\rm max}|} & \dots & 0 \\
    \vdots & &\ddots & \\
    0 & 0 &  \dots    & 1+\frac{\lambda_{N}}{|\lambda_{\rm max}|} 
    \end{bmatrix}V^{-1}
    \end{aligned}
\end{equation}

\textbf{All-pass.} An all-pass graph filter is simple:  $\mathcal{H}=I$.
The all-pass filter keeps all information of the degree signal and intuitively samples those nodes with large degrees.


\textbf{For correspondence graph.} When it comes to our filter, we first compute the generalized degree signal $s=[s_1,s_2,\dots,s_{N}]^T\in \mathbb{C}^{N\times1}$ where $s_i=\sum_j W_{\rm SOG_{ij}}$, and $N$ is the size of correspondence set. Then a high-pass graph filter is adopted to filter the high frequency information of $s$. For $G_{\rm corr}$, we define the high-pass graph filter as:
\begin{equation}
    \mathcal{H}={\rm Diag}(s)-W_{\rm SOG},
\end{equation}
or the Laplacian Matrix of the correspondence graph. In the graph vertex domain, the output for a signal $X=(x_i)$,  $(\mathcal{H}X)_i=s_ix_i-\sum_{j\in \mathcal{N}_i} W_{\rm SOG_{ij}}x_j$ reflects the difference between a node and the combination of its neighbors.


Then we have the response of signal $s$ corresponding to $\mathcal{H}$ as $f=\mathcal{H}s$. We further compute the response magnitude: $\pi_i=||f_i||_2^2 $
which quantifies the energy of the signal on each node after high-pass graph filtering. It reflects how much information we know about a signal value on the node from its neighbors in the graph.

\subsection{Stochastic Sampling}
\textbf{Sampling operator definition.} After obtaining the response magnitude of each node to the graph filter as shown in Fig.~\ref{fig:main-fig}, we perform sampling based on this response magnitude. Suppose we aim to sample $m$ components of a graph signal $x=\mathcal{H}s\in \mathbb{C}^n$ to produce a sampled signal $y=x_{\mathcal{M}}\in \mathbb{C}^m$, where $\mathcal{M}$ is the set of sampled indices. The sampling operator $\Psi$ is defined as a linear mapping from $\mathbb{C}^n$ to $\mathbb{C}^m$, $\Psi_{ij}=\delta_{j,\mathcal{M}_i}$ and the interpolation operator $\Phi$ is defined as a linear mapping from $\mathbb{C}^m$ to $\mathbb{C}^n$: 
\begin{subequations}
\begin{align}
y&=x_{\mathcal{M}}=\Psi x, \\
x'&=\Phi y=\Phi \Psi x,
\end{align}
\end{subequations}
where $x'\in \mathbb{C}^n$ is the recovery of the original signal. A properly designed sampling operator $\Psi$ aims to minimize the reconstruction error $||x-x'||$.

\textbf{Non-stochastic methods} attempt to create a well-designed deterministic sampling operator $\Psi$. \cite{chen2015discrete} finds the optimal sampling operator:
\begin{equation}
    \Psi_{\rm opt}=\arg\max_{\Psi}\sigma_{\rm min}(\Psi V_{\rm (K)}),
\end{equation}
where $\sigma_{\rm min}$ means the smallest singular value and $V_{\rm (K)}$ represents the independent columns in the eigenvectors $V$ of the graph shift $\mathcal{A}$. In practice, a greedy algorithm \cite{chen2015discrete} is used to find an approximate solution. It maintains $M$, a set of rows of $V_{\rm (K)}$, and loops to find another row $r$ in $V_{\rm (K)}$ to maximize $\sigma_{\rm min}$ of the matrix formed by $M+\{r\}$ until $|M|$ meets the termination condition. However, it is extremely slow when processing large matrices, as it involves a number of SVD decompositions
with a total complexity of $O(MN^3+M^3N)$ where $M$ is the sample size and $N$ is the original size. Proof will be given in \ref{complexity}.

\textbf{Stochastic sampling.} By contrast, we adopt a stochastic strategy. We consider $\pi_i$ fetched from \textit{Graph Filtering} as a sampling distribution and apply probability sampling on the initial correspondence set, resulting in a sampled set denoted as $C_{\rm sampled}$. $\pi_i$  approximates the sampling operator $\Psi$ and it is optimal in terms of minimizing the reconstruction error, according to proof in \cite{chen2017fast}, and is much faster, which will be proved in
\ref{stochastic}. 
A detailed proof of optimality is given in Appendix.A.5.


\section{Experiments}
 For information about datasets, evaluation metrics and implementation details, please refer to the Appendix A.6.
\subsection{Time-Accuracy Trade-off Comparison}
We perform an extensive comparison in Fig.~\ref{fig:teaser}. The correspondence based registration methods are presented for comparison.
All methods are tested on the KITTI dataset with FCGF as the correspondence generation descriptor.

Fig.~\ref{fig:teaser} demonstrates RR performance of different methods. \textbf{Our FastMAC can outperform all other methods even with a sample ratio as low as 5\%.} It runs nearly \textbf{80 times} faster than methods with comparable RR performance, and achieves a \textbf{40\% higher} RR when compared to methods that are almost as fast as it. Moreover, when sampling ratio declines to 20\%, our method runs at \textit{real-time} level, with a single registration requires less than 35ms. 


\subsection{Sampling Strategy Comparison}
\begin{figure}[t]
\centering
\includegraphics[width=1.00\linewidth]{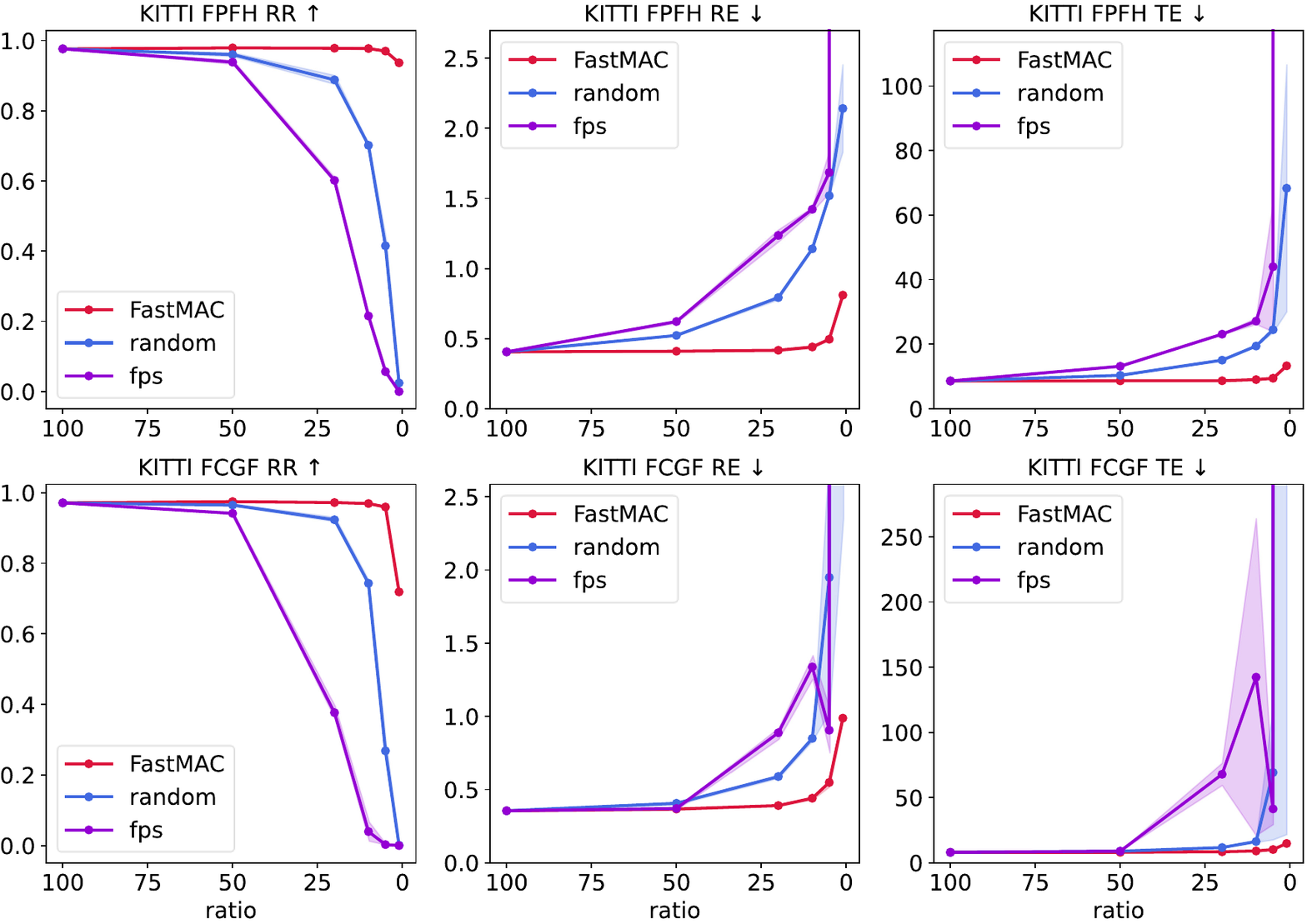}
\caption{
Sampling performance on KITTI. Each column represents a metric in TE,RE and RR and each row represents a setting composed of datasets and descriptors. Shaded areas represent variance from multiple runs.
}
\label{fig:down1}
\end{figure}
We also compare our method with different sampling strategies. This demonstrates that the MAC itself is still sensitive to the number of correspondences, thus showing that our method is superior and suited for Maximal Clique registration. The sampling strategies for comparison are Random Sampling and Furthest Point Sampling(FPS). Notably, a correspondence is not a traditional 3D point and we define their distance as the euclidean distance in 6D space. Both FPFH and FCGF descriptor are tested.



\textbf{Results on KITTI Dataset: } Fig.~\ref{fig:down1} shows results of FPFH and FCGF settings on KITTI Dataset. Our method maintains a consistent RR, RE and TE when sampling ratio declines from 100\% to 5\% and only becomes slightly worse at the 1\%. By comparison, Random Sampling and FPS strategy show rapid deterioration in performance. It's worth noticing that FPS behaves even worse than Random strategy, suggesting that a 6-dimensional vector space with Euclidean distance is ill-suited for the correspondence set. Another notable point is that the FPS performance shows dramatic fluctuations, indicating lack of robustness.


\begin{figure}[t]
\centering
\includegraphics[width=1.00\linewidth]{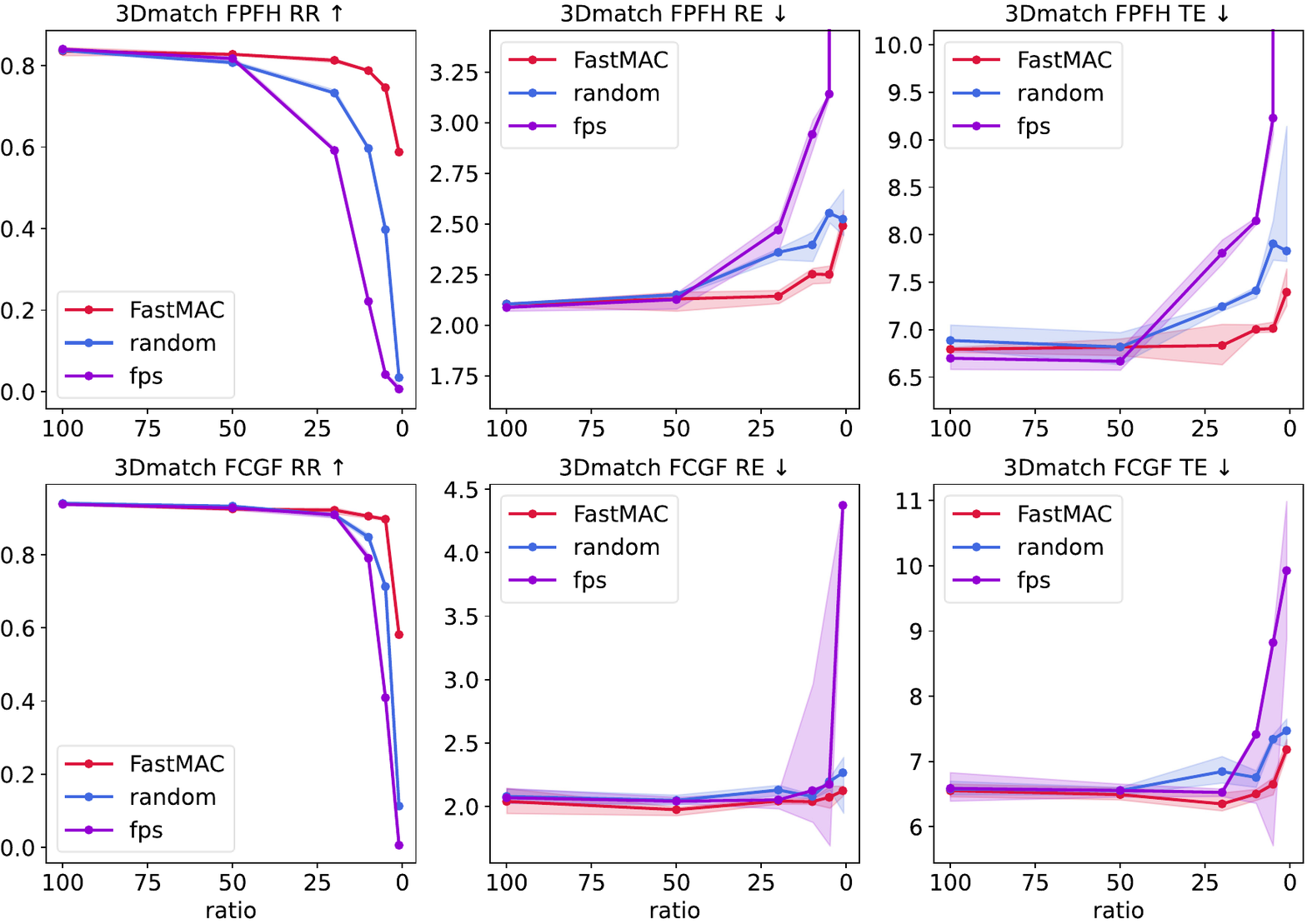}
\caption{
Sampling performance on 3DMatch. Each column represents a metric in TE,RE and RR and each row represents a setting composed of datasets and descriptors.
}
\label{fig:down2}
\end{figure}

\textbf{Results on 3DMatch Dataset: } As shown in Fig.~\ref{fig:down2}, our method still works well. Though the Random Strategy performs closely to our method on RE and TE, it has a much lower success rate. This means Random Strategy has a high performance on only a few point cloud pairs, which limits its usage in challenging real-world problems.




\begin{figure}[t]
\centering
\includegraphics[width=1.00\linewidth]{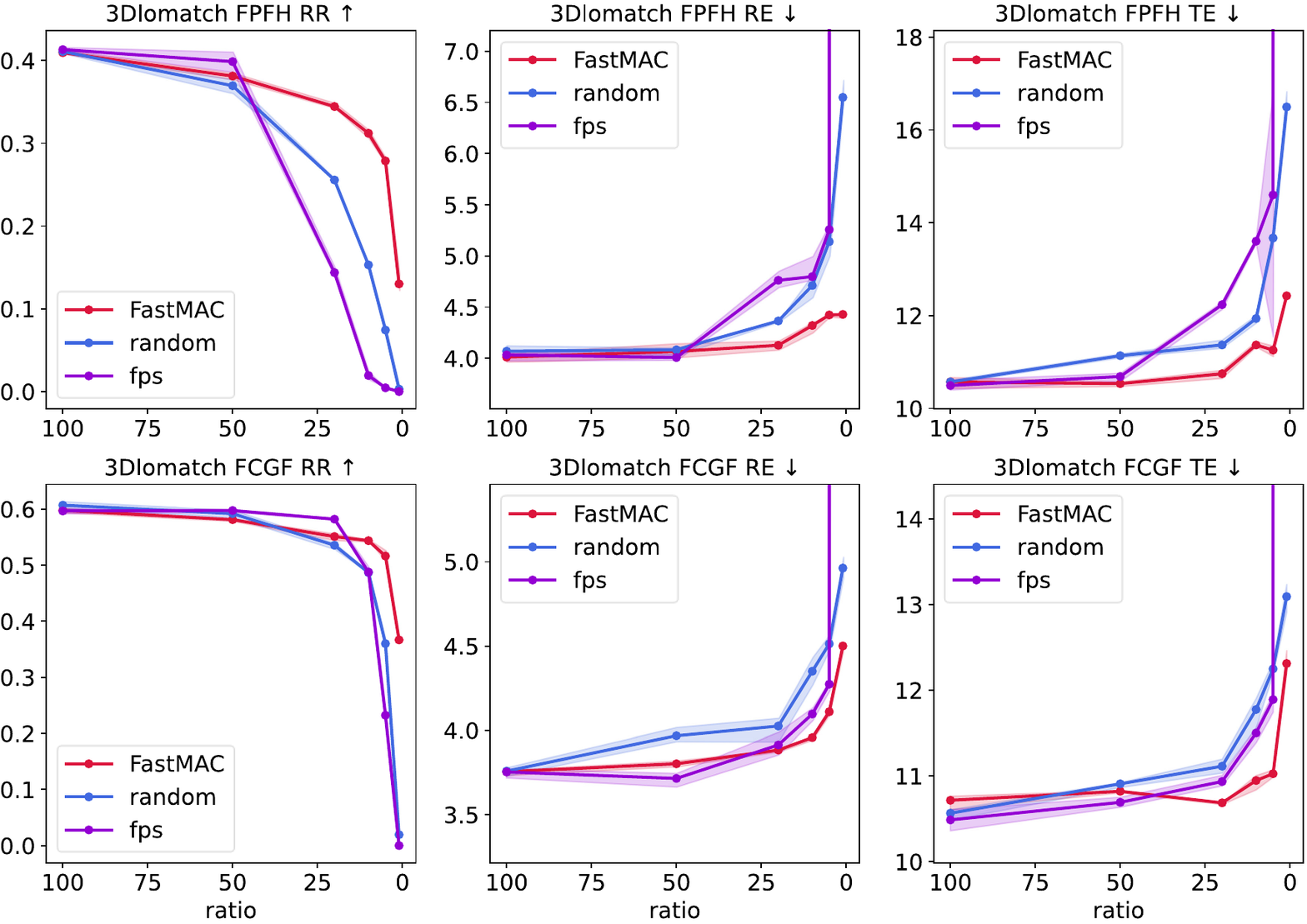}
\caption{
Sampling performance on 3DLoMatch. Each column represents a metric in TE,RE and RR and each row represents a setting composed of datasets and descriptors.
}
\label{fig:down3}
\end{figure}

\textbf{Results on 3DLoMatch Dataset: } As in Fig.~\ref{fig:down3}, RR of 3DLoMatch drops faster than 3DMatch when the sample ratio decreases for all three methods, due to low-overlap of this dataset. Still, our method significantly surpasses the performance of the other two methods, considering their almost-zero success rate at 1\% sample ratio.

\subsection{Time Profiling}

\begin{table}[t]
\centering
\resizebox{\linewidth}{!}{%
\begin{tabular}{c|c|ccccc|c}
\hline
method                   & ratio & \multicolumn{1}{c|}{Sampling} & \multicolumn{1}{c|}{GC} & \multicolumn{1}{c|}{MCS} & \multicolumn{1}{c|}{NCS} & \multicolumn{1}{c|}{PE} & Total \\ \hline
MAC                      & 100   & 0                               & 749.057                                & 61.390                                    & 132.809                                          & 18.056 &961.312       \\ \hline
\multirow{5}{*}{FastMAC} & 50    & 15.13                           & 129.404                                & 26.845                                    & 2.225                                          & 5.039 & 178.643        \\
                         & 20    & 15.02                           & 14.945                                & 4.794                                    & 0.408                                          & 0.903  & 36.070      \\
                         & 10    & 14.97                           & 3.369                                & 1.113                                    & 0.070                                          & 0.256   &19.778     \\
                         & 5     & 14.99                           & 0.971                                & 0.267                                    & 0.005                                          & 0.067   & 16.300     \\
                         & 1     & 15.06                           & 0.056                                & 0.001                                    & 0.001                                          & 0.019     &15.137   \\ \hline
\end{tabular}%
}
\caption{Time profiling for 3DMatch Dataset with FCGF descriptor. Vanilla MAC is compared with our FastMAC. Time consumption is measured in milliseconds. GC: Graph Construction; MCS: Maximal Clique Search; NCS: Node-guided Clique Selection; PE: Pose Estimation}
\label{tab:profile}
\end{table}

To demonstrate FastMAC's time efficiency, we study on-device time profiling to report time consumption. The original MAC is used for comparison. 

Fig.~\ref{fig:teaser} depicts our results on KITTI Dataset with FPFH descriptor. On the right side, we demonstrate time profiling of MAC and FastMAC with a 5\% sample ratio . For MAC, \textit{Graph Construction} and \textit{Maximal Cliques Search} occupy a major part of time consumption, pushing the total time to over \textit{1} second. Whilst for FastMAC, they are no longer bottlenecks. Middle shows variation of time taken versus the sample ratio. \textit{Sampling} gradually became the dominatant factor, with rest parts being barely time-consuming.

Table~\ref{tab:profile} further presents our findings on 3DMatch. Acceleration from \textit{Sampling} negates the time spent during the process itself. And its time decrease is essentially consistent across various sample rates, proving its efficiency.

\subsection{Comparison to State-of-the-arts}
\begin{table}[t]
\centering
\resizebox{\columnwidth}{!}{%
\begin{tabular}{c|ccc|ccc|c}
\hline
                                                                 & \multicolumn{3}{c|}{FPFH}               & \multicolumn{3}{c|}{FCGF}  &\multirow{2}{*}{Time(s)}                     \\
                                                                 & RR(\%)         & RE($^\circ$)  & TE(cm) & RR(\%)         & RE($^\circ$)  & TE(cm)        \\ \hline
FGR\cite{zhou2016fast}                                                              & 5.23           & 0.86          & 43.84  & 89.54          & 0.46          & 25.72   &    9.350  \\
TEASER++\cite{yang2020teaser}                                                         & 91.17          & 1.03          & 17.98  & 94.96          & 0.38          & 13.69  &      0.070 \\
RANSAC-4M\cite{fischler1981random}                                                        & 74.41          & 1.55          & 30.20  & 80.36          & 0.73          & 26.79     &  52.40  \\
CG-SAC\cite{yang2021sac}                                                        & 74.23          & 0.73          & 14.02  & 83.24          & 0.56          & 22.96       & 2.140 \\
SC2-PCR\cite{chen2022sc2}                                                          & 96.40          & \textbf{0.41} & 8.00   & 97.12          & 0.41          & 9.71      & 0.850   \\
DGR\cite{choy2020deep}                                                              & 77.12          & 1.64          & 33.10  & 94.90          & 0.34          & 21.70   &   0.330   \\
PointDSC\cite{bai2021pointdsc}                                                         & 96.40          & 0.38          & 8.35   & 96.40          & 0.61          & 13.42   & 0.130     \\
MAC\cite{zhang20233d}                                                              & 97.66          & \textbf{0.41} & 8.61   & 97.25          & \textbf{0.36} & {\ul 8.00} &0.570   \\ \hline
FastMAC@50                                                       & {\ul 97.84}    & \textbf{0.41} & 8.61   & \textbf{97.84} & \textbf{0.36} & \textbf{7.98} &0.114 \\
\begin{tabular}[c]{@{}c@{}}FastMAC@20\\ (\textbf{real-time})\end{tabular} & \textbf{98.02} & \textbf{0.41} & 8.64   & {\ul 97.48}    & {\ul 0.38}    & 8.20    &  0.028    \\ \hline
\end{tabular}%
}
\caption{Comparison with baseline methods on KITTI Dataset. The best and second-to-best results of baseline methods are repectively marked in bold and underlined. FastMAC@$\rm x$ refers to our method sampling at $\rm x\%$ ratio.}
\label{tab:kitti}
\end{table}

\begin{table}[t]
\centering
\resizebox{\columnwidth}{!}{%
\begin{tabular}{c|ccc|ccc|c}
\hline
\multirow{2}{*}{} & \multicolumn{3}{c|}{FPFH}                      & \multicolumn{3}{c|}{FCGF} & \multirow{2}{*}{Time(s)}                      \\
                  & RR(\%)         & RE($^\circ$)  & TE(cm)        & RR(\%)         & RE($^\circ$)  & TE(cm) &       \\ \hline
RANSAC-1M\cite{fischler1981random}         & 64.20          & 4.05          & 11.35         & 88.42          & 3.05          & 9.42    &   23.30   \\
RANSAC-4M\cite{fischler1981random}         & 66.10          & 3.95          & 11.03         & 91.44          & 2.69          & 8.38   &  95.80     \\
GC-RANSAC\cite{barath2018graph}         & 67.65          & 2.33          & 6.87          & 92.05          & 2.33          & 7.11    &  0.450    \\
TEASER++\cite{yang2020teaser}          & 75.48          & 2.48          & 7.31          & 85.77          & 2.73          & 8.66     & 0.030    \\
SC2-PCR\cite{chen2022sc2}           & {\ul 83.73}    & {\ul 2.18}    & {\ul 6.70}    & {\ul 93.16}    & {\ul 2.09}    & \textbf{6.51}  &0.920 \\
3DRegNet\cite{pais20203dregnet}          & 26.31          & 3.75          & 9.60          & 77.76          & 2.74          & 8.13      &  0.050  \\
DGR\cite{choy2020deep}               & 32.84          & 2.45          & 7.53          & 88.85          & 2.28          & 7.02       & 1.260  \\
PointDSC\cite{bai2021pointdsc}          & 72.95          & {\ul 2.18}    & \textbf{6.45} & 91.87          & 2.10          & 6.54     & 0.140    \\
MAC\cite{zhang20233d}               & \textbf{83.90} & \textbf{2.11} & 6.80          & \textbf{93.72} & \textbf{2.02} & {\ul 6.54} & 0.914  \\ \hline
FastMAC@50        & 82.87          & 2.15          & 6.73          & 92.67          & 2.00          & 6.47      &  0.203  \\
\begin{tabular}[c]{@{}c@{}}FastMAC@20\\ (\textbf{real-time})\end{tabular}       & 80.71          & 2.17          & 6.81          & 92.30          & 2.02          & 6.52    &  0.038    \\
\hline
\end{tabular}%
}
\caption{Comparison with baseline methods on 3DMatch Dataset. The best and second-to-best results of baseline methods are repectively marked in bold and underlined. FastMAC@$
\rm x$ refers to our method sampling at $\rm x\%$ ratio.}
\label{tab:3dmatch}
\end{table}

\begin{table}[t]
\centering
\resizebox{\columnwidth}{!}{%
\begin{tabular}{c|ccc|ccc|c}
\hline
           & \multicolumn{3}{c|}{FPFH}                      & \multicolumn{3}{c|}{FCGF}       & \multirow{2}{*}{Time(s)}                \\ 
           & RR(\%)             & RE($^\circ$)            & TE(cm)            & RR(\%)             & RE($^\circ$)            & TE(cm)            \\ \hline
RANSAC-1M\cite{fischler1981random}  & 0.67           & 10.27         & 15.06         & 9.77           & 7.01          & 14.87     & 19.60   \\
RANSAC-4M\cite{fischler1981random}  & 0.45           & 10.39         & 20.03         & 10.44          & 6.91          & 15.14    &  86.30   \\
TEASER++\cite{yang2020teaser}   & 35.15          & 4.38          & 10.96         & 46.76          & 4.12          & 12.89   &  0.030    \\
SC2-PCR\cite{chen2022sc2}    & {\ul 38.57}    & {\ul 4.03}    & 10.31         & {\ul 58.73}    & {\ul 3.80}    & {\ul 10.44}  & 0.720\\
DGR\cite{choy2020deep}        & 19.88          & 5.07          & 13.53         & 43.80          & 4.17          & 10.82    &  1.220   \\
PointDSC\cite{bai2021pointdsc}   & 20.38          & 4.04          & {\ul 10.25}   & 56.20          & 3.87          & 10.48     & 0.140   \\
MAC\cite{zhang20233d}        & \textbf{40.88} & \textbf{3.66} & \textbf{9.45} & \textbf{59.85} & \textbf{3.50} & \textbf{9.75}  & 1.181\\ \hline
FastMAC@50 & 38.46          & 4.04          & 10.47         & 58.23          & 3.80          & 10.81    & 0.271    \\
\begin{tabular}[c]{@{}c@{}}FastMAC@20\\ (\textbf{real-time})\end{tabular} & 34.31          & 4.12          & 10.82         & 55.25          & 3.84          & 10.71   &   0.051   \\
\hline
\end{tabular}%
}
\caption{Comparison with baseline methods on 3DLoMatch Dataset. The best and second-to-best results of baseline methods are repectively marked in bold and underlined.}
\label{tab:3dlomatch}
\end{table}

Our method is compared with baseline approaches on the 3DMatch, 3DLomatch, and KITTI datasets, and the outcomes can be found in Tables~\ref{tab:kitti},~\ref{tab:3dmatch} and~\ref{tab:3dlomatch}. When sampling at various ratios, FastMAC reports no significant decrease in performance, remaining competitive with other state-of-the-art methods. This showcases the efficacy of our sampling technique, which accelerates the state-of-the-art MAC method whilst maintaining accuracy.

\subsection{Descriptor Robustness}

\begin{figure}[t]
\centering
\includegraphics[width=1.00\linewidth]{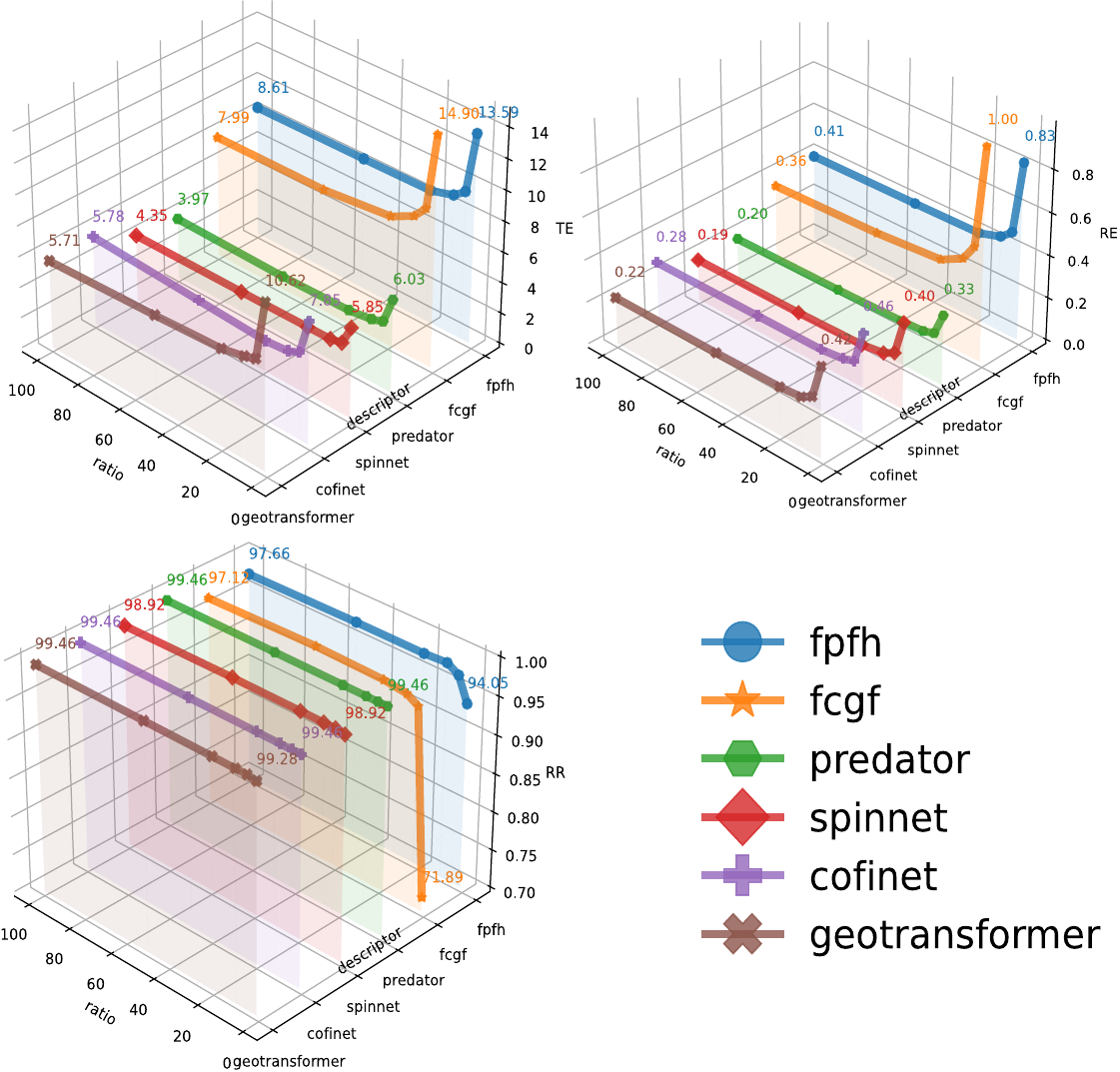}
\caption{
Performance on KITTI with various descriptors. 
}
\label{fig:desc}
\end{figure}

Since our method accepts correspondences as input, it is crucial to demonstrate its ability to work with correspondences generated by different descriptors.
We perform extensive experiments with various descriptors, including FPFH\cite{rusu2009fast}, FCGF\cite{choy2019fully}, Predator\cite{huang2021predator}, Spinnet\cite{ao2021spinnet}, Cofinet\cite{yu2021cofinet} and Geotransformer\cite{qin2022geometric}. These descriptors are used to generate point-wise features which are subsequently used to obtain correspondences.  We adopt KITTI as the dataset and the results are presented in Fig.~\ref{fig:desc}.

For RR metric, stronger descriptors like Geotransformer, Cofinet, Spinnet and Predator exhibit amazing excellence, with their RR remaining unaffected by the sample ratio, while FCGF and FPFH perform slightly worse when sample ratio comes to 1\%. For RE and TE, most descriptors behave similarly. These metrics first stabilize and then slowly increase as the sampling rate decreases. It can therefore be concluded that our method is highly robust to the correspondences produced by different descriptors.
\section{Ablation Study}
In this section, we mainly focus on the analysis of the core parts of our method, i.e, \textit{High-pass Graph Filtering} and \textit{Stochastically Sampling}. 
Since $\rm xyz$ coordinates of the point cloud can also be a graph signal and has been widely used before \cite{chen2017fast}, we further compare $\rm xyz$ signal with our generalized degree signal to demonstrate our superiority. 

\subsection{High-pass, Low-pass and All-pass Filter}
In this part, we compare our high-pass filter with a low-pass filter and an all-pass filter to demonstrate its efficacy. The filters are implemented as described in \ref{methods}.
For the Haar-like low-pass filter, we choose the graph shift \(\mathcal{A}\) to be \(D^{-1}W\) where \(D\) is the generalized degree matrix and \(W\) is the adjacency matrix. Then we have the sample distribution 
$\pi_i=||((I+D^{-1}W)s)_i||_2$
in which $s$ is the generalized degree signal. For the all-pass filter, the corresponding sampling strategy is $\pi_i=||s_i||_2$.



\begin{table}[t]
\centering
\renewcommand\arraystretch{1.5}
\resizebox{\columnwidth}{!}{%
\begin{tabular}{c|ccc|ccc|ccc}
\hline
\multirow{2}{*}{Ratio} & \multicolumn{3}{c|}{High-pass}                    & \multicolumn{3}{c|}{All-pass}  & \multicolumn{3}{c}{Low-pass}    \\ \cline{2-10} 
                       & RR(\%)         & RE ($^\circ$)   & TE(cm)          & RR(\%) & RE ($^\circ$) & TE(cm) & RR(\%) & RE ($^\circ$) & TE(cm)  \\ \hline
50                     & \textbf{97.66} & \textbf{0.368} & \textbf{8.016}  & 97.12  & 0.368        & 8.098  & 97.48  & 0.368        & 8.135   \\
20                     & \textbf{97.66} & \textbf{0.391} & \textbf{8.457}  & 96.58  & 0.395        & 8.672  & 97.30  & 0.403        & 8.648   \\
10                     & \textbf{96.94} & \textbf{0.446} & \textbf{9.201}  & 96.94  & 0.480        & 9.329  & 94.96  & 0.468        & 9.417   \\
5                      & \textbf{96.04} & \textbf{0.525} & \textbf{10.038} & 93.33  & 0.583        & 10.907 & 90.09  & 0.616        & 11.182  \\
1                      & \textbf{71.89} & \textbf{0.997} & \textbf{14.899} & 28.11  & 1.058        & 15.217 & 13.33  & 3.516        & 106.211 \\ \hline
\end{tabular}%
}
\caption{Registration results on KITTI FCGF dataset for comparison between the High-pass, Low-pass and All-pass filters.}
\label{tab:filter}
\end{table}

We use these types of filters to sample the correspondences and feed the output into the MAC module. The result is shown in Table.~\ref{tab:filter}. The high-pass filter is more effective , providing evidence to support our intuition. In fact, an all-pass filter solely samples out nodes with high degrees, whereas a low-pass filter samples nodes which lie within a clique. If the clique is big in size, low-pass and all-pass filters function in a manner similar to random sampling. That is why these two filters do not function well.

\subsection{Stochastic v.s. Non-stochastic}\label{stochastic}


To clarify the efficiency of our stochastic method, we will give both theoretical analysis and experimental analysis. 

\paragraph{Theoretical Analysis}\label{complexity}
Since both stochastic and non-stochastic methods involve computing the signal after filtering, we only analyze operations after the computation. Suppose we are sampling $M$ nodes from the original $N$ nodes.

     {\textbf{Non-stochastic:}}  $V_{\rm (K)}$ is first required for the graph shift $\mathcal{A}$, which is $O(N^3)$ complexity \cite{pan1999complexity}. Then we enter the greedy algorithm which loops for $M$ times. In loop $i$, $N-i$ times of computing the smallest singular value is performed, each using $O(N^3+(i+1)^2N)$ \cite{holmes2007fast}. After summing them all, we get the final computation complexity
\begin{equation}
	\begin{split}
	&O(N^3)+\sum_{i=0}^{M-1}O(N^3+(i+1)^2N)\\
        =&O(MN^3+M^3N)
	\end{split}
\end{equation}

{\textbf{Stochastic:}} Our stochastic method simply use the norm of the filtered signal $y$ as a distribution and samples $M$ times from it. The time complexity is $O(M)+O(N)$.

\paragraph{Experimental Analysis}

\begin{figure}[t]
\centering
\includegraphics[width=1.00\linewidth]{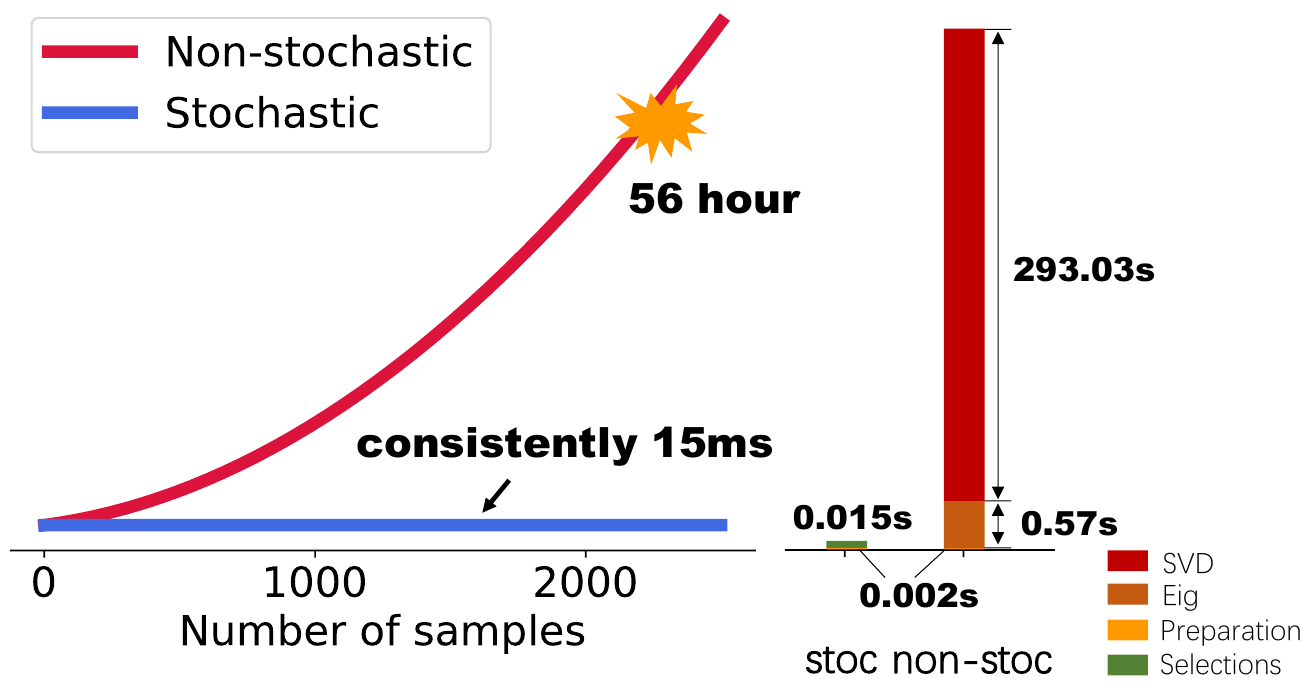}
\caption{
Left figure shows how time increases with number of samples on KITTI FCGF dataset. Right is time profiling when sampling 50 from 5000 correspondences. \textbf{Preparation}: Computation of the graph shift and the degree score. \textbf{Sampling}: For stochastic method, only 50 selection operations are required according to the given distribution. For non-stochastic method, it includes an eigendecomposition and 50 loops of $\sim$5000-scale SVD.
}
\label{fig:stochastic}
\end{figure}

We present our experimental results in Fig.~\ref{fig:stochastic}. The time taken by the stochastic method remains consistently stable at 15 ms as the sample rate increases. By contrast, the non-stochastic method's time consumption increases in a polynomial trend, necessitating two days to obtain 2500 samples! This is largely caused by the frequent, multiple large matrix SVD decompositions.

\subsection{XYZ Signal v.s. Generalized Degree Signal}
In this section, we compare our generalized degree signal with xyz signal which are commonly used in point cloud sampling. As we accept correspondences as input, we denote the source point cloud formed by source points as $Q_{\rm s}\in\mathbb{C}^{N\times3}$ and target point cloud $Q_{\rm t}\in\mathbb{C}^{N\times3}$. They are considered as graph signals and pass through a high-pass filter which is created by KNN adjacency matrix following \cite{chen2017fast} to detect their contour points. The response are denoted as $q_{\rm s}\in \mathbb{C}^{N\times3}$ and $q_{\rm t}\in\mathbb{C}^{N\times3}$. The response magnitude $||q_{\rm s}||_2\in\mathbb{C}^{N}$ and $||q_{\rm t}||_2\in\mathbb{C}^{N}$ are then scaled to 0-1 to form the sampling distribution $\pi_1$, $\pi_2$. 

\begin{figure}[t]
\centering
\includegraphics[width=1.00\linewidth]{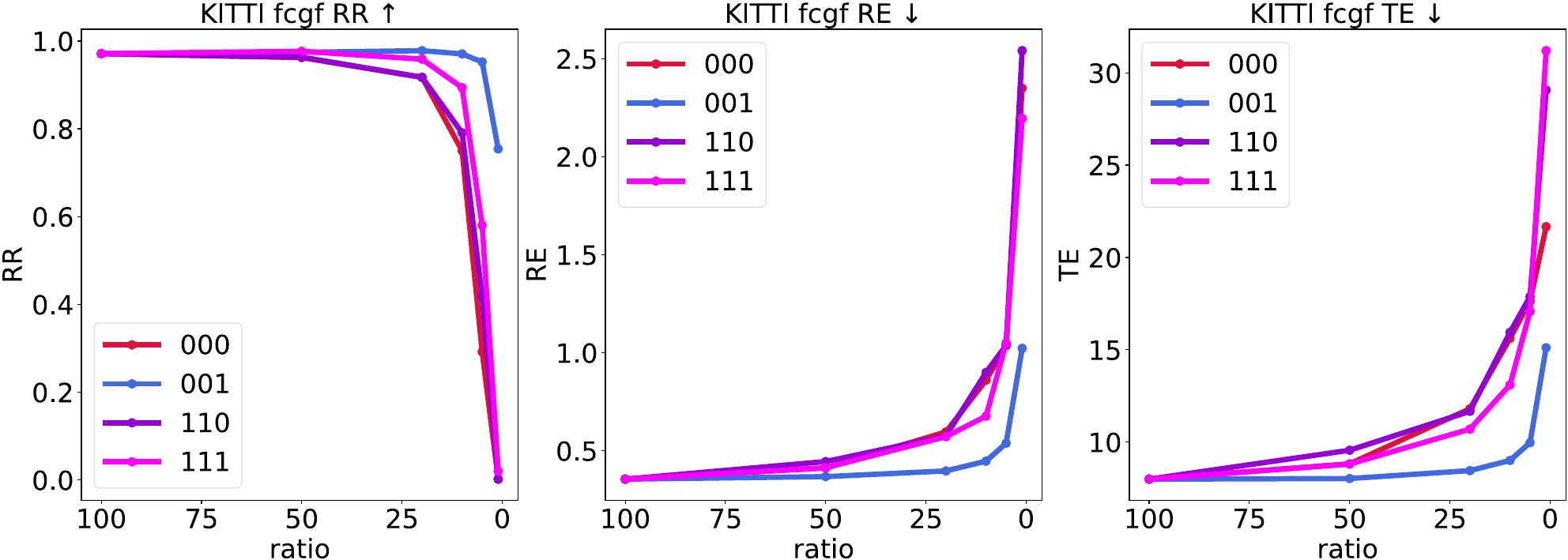}
\caption{
Sampling performance on KITTI FCGF with different degree signal settings. \textbf{000}: random sampling.
    \textbf{110}: sample guided by xyz signals from source and target pointcloud.
    \textbf{001}: sample guided by our generalized degree signal.
    \textbf{111}: sample guided by simply adding both signals.
}
\label{fig:ablation}
\end{figure}

Fig.~\ref{fig:ablation} demonstrates the results of four settings. 
The signal of the generalized degree performs the best, while xyz signal does not differ significantly from random sampling. When combined with xyz signal, the performance of generalized degree signal gets worse, indicating that the information within the generalized degree signal is contaminated by xyz signal.


\section{Conclusion} \label{sec9}
In this paper, we propose a stochastic spectral correspondence graph sampling method to discover high-frequency nodes, boosting Maximal Clique Registration to real-time level with little performance loss. Moreover, it's robust to various descriptors, showing its potential for usage in real-time complex applications. Still we have limitations. As shown in Fig.~\ref{fig:teaser}, when sample ratio decreases, graph construction gradually becomes a bottleneck. In the future we plan to solve this issue by learning graph prior without building a graph, thus eliminating this cost. 

{
\small
\bibliographystyle{ieeenat_fullname}
\bibliography{main}

\begin{thebibliography}{76}
\providecommand{\natexlab}[1]{#1}
\providecommand{\url}[1]{\texttt{#1}}
\expandafter\ifx\csname urlstyle\endcsname\relax
  \providecommand{\doi}[1]{doi: #1}\else
  \providecommand{\doi}{doi: \begingroup \urlstyle{rm}\Url}\fi

\bibitem[Ao et~al.(2021)Ao, Hu, Yang, Markham, and Guo]{ao2021spinnet}
Sheng Ao, Qingyong Hu, Bo Yang, Andrew Markham, and Yulan Guo.
\newblock Spinnet: Learning a general surface descriptor for 3d point cloud registration.
\newblock In \emph{Proceedings of the IEEE/CVF conference on computer vision and pattern recognition}, pages 11753--11762, 2021.

\bibitem[Bai et~al.(2021)Bai, Luo, Zhou, Chen, Li, Hu, Fu, and Tai]{bai2021pointdsc}
Xuyang Bai, Zixin Luo, Lei Zhou, Hongkai Chen, Lei Li, Zeyu Hu, Hongbo Fu, and Chiew-Lan Tai.
\newblock Pointdsc: Robust point cloud registration using deep spatial consistency.
\newblock In \emph{Proceedings of the IEEE/CVF Conference on Computer Vision and Pattern Recognition}, pages 15859--15869, 2021.

\bibitem[Barath and Matas(2018)]{barath2018graph}
Daniel Barath and Ji{\v{r}}{\'\i} Matas.
\newblock Graph-cut ransac.
\newblock In \emph{Proceedings of the IEEE conference on computer vision and pattern recognition}, pages 6733--6741, 2018.

\bibitem[Bazin et~al.(2013)Bazin, Seo, and Pollefeys]{bazin2013globally}
Jean-Charles Bazin, Yongduek Seo, and Marc Pollefeys.
\newblock Globally optimal consensus set maximization through rotation search.
\newblock In \emph{Computer Vision--ACCV 2012: 11th Asian Conference on Computer Vision, Daejeon, Korea, November 5-9, 2012, Revised Selected Papers, Part II 11}, pages 539--551. Springer, 2013.

\bibitem[Besl and McKay(1992)]{besl1992method}
Paul~J Besl and Neil~D McKay.
\newblock Method for registration of 3-d shapes.
\newblock In \emph{Sensor fusion IV: control paradigms and data structures}, pages 586--606. Spie, 1992.

\bibitem[Breuel(2003)]{breuel2003implementation}
Thomas~M Breuel.
\newblock Implementation techniques for geometric branch-and-bound matching methods.
\newblock \emph{Computer Vision and Image Understanding}, 90\penalty0 (3):\penalty0 258--294, 2003.

\bibitem[Bustos and Chin(2017)]{bustos2017guaranteed}
Alvaro~Parra Bustos and Tat-Jun Chin.
\newblock Guaranteed outlier removal for point cloud registration with correspondences.
\newblock \emph{IEEE transactions on pattern analysis and machine intelligence}, 40\penalty0 (12):\penalty0 2868--2882, 2017.

\bibitem[Cao et~al.(2023)Cao, Zhu, Ren, Choset, and Zhang]{cao2023representation}
C Cao, H Zhu, Z Ren, H Choset, and J Zhang.
\newblock Representation granularity enables time-efficient autonomous exploration in large, complex worlds.
\newblock \emph{Science Robotics}, 8\penalty0 (80):\penalty0 eadf0970, 2023.

\bibitem[Carlone(2022)]{carlone2022estimation}
Luca Carlone.
\newblock Estimation contracts for outlier-robust geometric perception.
\newblock \emph{arXiv preprint arXiv:2208.10521}, 2022.

\bibitem[Chen et~al.(2015)Chen, Varma, Sandryhaila, and Kova{\v{c}}evi{\'c}]{chen2015discrete}
Siheng Chen, Rohan Varma, Aliaksei Sandryhaila, and Jelena Kova{\v{c}}evi{\'c}.
\newblock Discrete signal processing on graphs: Sampling theory.
\newblock \emph{IEEE transactions on signal processing}, 63\penalty0 (24):\penalty0 6510--6523, 2015.

\bibitem[Chen et~al.(2017)Chen, Tian, Feng, Vetro, and Kova{\v{c}}evi{\'c}]{chen2017fast}
Siheng Chen, Dong Tian, Chen Feng, Anthony Vetro, and Jelena Kova{\v{c}}evi{\'c}.
\newblock Fast resampling of three-dimensional point clouds via graphs.
\newblock \emph{IEEE Transactions on Signal Processing}, 66\penalty0 (3):\penalty0 666--681, 2017.

\bibitem[Chen et~al.(2022{\natexlab{a}})Chen, Zhao, Zhou, and Zhang]{chen2022pq}
Xiaoxue Chen, Hao Zhao, Guyue Zhou, and Ya-Qin Zhang.
\newblock Pq-transformer: Jointly parsing 3d objects and layouts from point clouds.
\newblock \emph{IEEE Robotics and Automation Letters}, 7\penalty0 (2):\penalty0 2519--2526, 2022{\natexlab{a}}.

\bibitem[Chen et~al.(2022{\natexlab{b}})Chen, Sun, Yang, and Tao]{chen2022sc2}
Zhi Chen, Kun Sun, Fan Yang, and Wenbing Tao.
\newblock Sc2-pcr: A second order spatial compatibility for efficient and robust point cloud registration.
\newblock In \emph{Proceedings of the IEEE/CVF Conference on Computer Vision and Pattern Recognition}, pages 13221--13231, 2022{\natexlab{b}}.

\bibitem[Chetverikov et~al.(2005)Chetverikov, Stepanov, and Krsek]{chetverikov2005robust}
Dmitry Chetverikov, Dmitry Stepanov, and Pavel Krsek.
\newblock Robust euclidean alignment of 3d point sets: the trimmed iterative closest point algorithm.
\newblock \emph{Image and vision computing}, 23\penalty0 (3):\penalty0 299--309, 2005.

\bibitem[Chin and Suter(2022)]{chin2022maximum}
Tat-Jun Chin and David Suter.
\newblock \emph{The maximum consensus problem: recent algorithmic advances}.
\newblock Springer Nature, 2022.

\bibitem[Chin et~al.(2018)Chin, Cai, and Neumann]{chin2018robust}
Tat-Jun Chin, Zhipeng Cai, and Frank Neumann.
\newblock Robust fitting in computer vision: Easy or hard?
\newblock In \emph{Proceedings of the European Conference on Computer Vision (ECCV)}, pages 701--716, 2018.

\bibitem[Choy et~al.(2019)Choy, Park, and Koltun]{choy2019fully}
Christopher Choy, Jaesik Park, and Vladlen Koltun.
\newblock Fully convolutional geometric features.
\newblock In \emph{Proceedings of the IEEE/CVF international conference on computer vision}, pages 8958--8966, 2019.

\bibitem[Choy et~al.(2020)Choy, Dong, and Koltun]{choy2020deep}
Christopher Choy, Wei Dong, and Vladlen Koltun.
\newblock Deep global registration.
\newblock In \emph{Proceedings of the IEEE/CVF conference on computer vision and pattern recognition}, pages 2514--2523, 2020.

\bibitem[Chum and Matas(2005)]{chum2005matching}
Ondrej Chum and Jiri Matas.
\newblock Matching with prosac-progressive sample consensus.
\newblock In \emph{2005 IEEE computer society conference on computer vision and pattern recognition (CVPR'05)}, pages 220--226. IEEE, 2005.

\bibitem[Curless and Levoy(1996)]{curless1996volumetric}
Brian Curless and Marc Levoy.
\newblock A volumetric method for building complex models from range images.
\newblock In \emph{Proceedings of the 23rd annual conference on Computer graphics and interactive techniques}, pages 303--312, 1996.

\bibitem[Delmerico et~al.(2018)Delmerico, Isler, Sabzevari, and Scaramuzza]{delmerico2018comparison}
Jeffrey Delmerico, Stefan Isler, Reza Sabzevari, and Davide Scaramuzza.
\newblock A comparison of volumetric information gain metrics for active 3d object reconstruction.
\newblock \emph{Autonomous Robots}, 42\penalty0 (2):\penalty0 197--208, 2018.

\bibitem[Enqvist et~al.(2009)Enqvist, Josephson, and Kahl]{enqvist2009optimal}
Olof Enqvist, Klas Josephson, and Fredrik Kahl.
\newblock Optimal correspondences from pairwise constraints.
\newblock In \emph{2009 IEEE 12th international conference on computer vision}, pages 1295--1302. IEEE, 2009.

\bibitem[Fischler and Bolles(1981)]{fischler1981random}
Martin~A Fischler and Robert~C Bolles.
\newblock Random sample consensus: a paradigm for model fitting with applications to image analysis and automated cartography.
\newblock \emph{Communications of the ACM}, 24\penalty0 (6):\penalty0 381--395, 1981.

\bibitem[Gao et~al.(2023{\natexlab{a}})Gao, Tian, Li, Chen, Zhao, Zhou, Chen, and Zha]{gao2023semi}
Huan-ang Gao, Beiwen Tian, Pengfei Li, Xiaoxue Chen, Hao Zhao, Guyue Zhou, Yurong Chen, and Hongbin Zha.
\newblock From semi-supervised to omni-supervised room layout estimation using point clouds.
\newblock In \emph{2023 IEEE International Conference on Robotics and Automation (ICRA)}, pages 2803--2810. IEEE, 2023{\natexlab{a}}.

\bibitem[Gao et~al.(2023{\natexlab{b}})Gao, Tian, Li, Zhao, and Zhou]{gao2023dqs3d}
Huan-ang Gao, Beiwen Tian, Pengfei Li, Hao Zhao, and Guyue Zhou.
\newblock Dqs3d: Densely-matched quantization-aware semi-supervised 3d detection.
\newblock In \emph{Proceedings of the IEEE/CVF International Conference on Computer Vision}, pages 21905--21915, 2023{\natexlab{b}}.

\bibitem[Geiger et~al.(2012)Geiger, Lenz, and Urtasun]{geiger2012we}
Andreas Geiger, Philip Lenz, and Raquel Urtasun.
\newblock Are we ready for autonomous driving? the kitti vision benchmark suite.
\newblock In \emph{2012 IEEE conference on computer vision and pattern recognition}, pages 3354--3361. IEEE, 2012.

\bibitem[Granger and Pennec(2002)]{granger2002multi}
S{\'e}bastien Granger and Xavier Pennec.
\newblock Multi-scale em-icp: A fast and robust approach for surface registration.
\newblock In \emph{Computer Vision—ECCV 2002: 7th European Conference on Computer Vision Copenhagen, Denmark, May 28--31, 2002 Proceedings, Part IV 7}, pages 418--432. Springer, 2002.

\bibitem[Hartley and Kahl(2009)]{hartley2009global}
Richard~I Hartley and Fredrik Kahl.
\newblock Global optimization through rotation space search.
\newblock \emph{International Journal of Computer Vision}, 82\penalty0 (1):\penalty0 64--79, 2009.

\bibitem[Holmes et~al.(2007)Holmes, Gray, and Isbell]{holmes2007fast}
Michael Holmes, Alexander Gray, and Charles Isbell.
\newblock Fast svd for large-scale matrices.
\newblock In \emph{Workshop on Efficient Machine Learning at NIPS}, pages 249--252, 2007.

\bibitem[Huang et~al.(2021)Huang, Gojcic, Usvyatsov, Wieser, and Schindler]{huang2021predator}
Shengyu Huang, Zan Gojcic, Mikhail Usvyatsov, Andreas Wieser, and Konrad Schindler.
\newblock Predator: Registration of 3d point clouds with low overlap.
\newblock In \emph{Proceedings of the IEEE/CVF Conference on computer vision and pattern recognition}, pages 4267--4276, 2021.

\bibitem[Kaneko et~al.(2003)Kaneko, Kondo, and Miyamoto]{kaneko2003robust}
Shun'ichi Kaneko, Tomonori Kondo, and Atsushi Miyamoto.
\newblock Robust matching of 3d contours using iterative closest point algorithm improved by m-estimation.
\newblock \emph{Pattern Recognition}, 36\penalty0 (9):\penalty0 2041--2047, 2003.

\bibitem[Konouchine et~al.(2005)Konouchine, Gaganov, and Veznevets]{konouchine2005amlesac}
Anton Konouchine, Victor Gaganov, and Vladimir Veznevets.
\newblock Amlesac: A new maximum likelihood robust estimator.
\newblock \emph{Proc. of Graphicon-2005. Novosibirsk}, pages 93--100, 2005.

\bibitem[Leordeanu and Hebert(2005)]{leordeanu2005spectral}
Marius Leordeanu and Martial Hebert.
\newblock A spectral technique for correspondence problems using pairwise constraints.
\newblock In \emph{Tenth IEEE International Conference on Computer Vision (ICCV'05) Volume 1}, pages 1482--1489. IEEE, 2005.

\bibitem[Li(2009)]{li2009consensus}
Hongdong Li.
\newblock Consensus set maximization with guaranteed global optimality for robust geometry estimation.
\newblock In \emph{2009 IEEE 12th International Conference on Computer Vision}, pages 1074--1080. IEEE, 2009.

\bibitem[Li et~al.(2023)Li, Zhao, Shi, Zhao, Yuan, Zhou, and Zhang]{li2023lode}
Pengfei Li, Ruowen Zhao, Yongliang Shi, Hao Zhao, Jirui Yuan, Guyue Zhou, and Ya-Qin Zhang.
\newblock Lode: Locally conditioned eikonal implicit scene completion from sparse lidar.
\newblock In \emph{2023 IEEE International Conference on Robotics and Automation (ICRA)}, pages 8269--8276. IEEE, 2023.

\bibitem[Lin et~al.(2022)Lin, Lin, and Wang]{lin2022k}
Yu-Kai Lin, Wen-Chieh Lin, and Chieh-Chih Wang.
\newblock K-closest points and maximum clique pruning for efficient and effective 3-d laser scan matching.
\newblock \emph{IEEE Robotics and Automation Letters}, 7\penalty0 (2):\penalty0 1471--1477, 2022.

\bibitem[Liu et~al.(2010)Liu, Yuen, and Torralba]{liu2010sift}
Ce Liu, Jenny Yuen, and Antonio Torralba.
\newblock Sift flow: Dense correspondence across scenes and its applications.
\newblock \emph{IEEE transactions on pattern analysis and machine intelligence}, 33\penalty0 (5):\penalty0 978--994, 2010.

\bibitem[Lusk et~al.(2021)Lusk, Fathian, and How]{lusk2021clipper}
Parker~C Lusk, Kaveh Fathian, and Jonathan~P How.
\newblock Clipper: A graph-theoretic framework for robust data association.
\newblock In \emph{2021 IEEE International Conference on Robotics and Automation (ICRA)}, pages 13828--13834. IEEE, 2021.

\bibitem[Maier-Hein et~al.(2011)Maier-Hein, Franz, Dos~Santos, Schmidt, Fangerau, Meinzer, and Fitzpatrick]{maier2011convergent}
Lena Maier-Hein, Alfred~Michael Franz, Thiago~R Dos~Santos, Mirko Schmidt, Markus Fangerau, Hans-Peter Meinzer, and J~Michael Fitzpatrick.
\newblock Convergent iterative closest-point algorithm to accomodate anisotropic and inhomogenous localization error.
\newblock \emph{IEEE transactions on pattern analysis and machine intelligence}, 34\penalty0 (8):\penalty0 1520--1532, 2011.

\bibitem[Mangelson et~al.(2018)Mangelson, Dominic, Eustice, and Vasudevan]{mangelson2018pairwise}
Joshua~G Mangelson, Derrick Dominic, Ryan~M Eustice, and Ram Vasudevan.
\newblock Pairwise consistent measurement set maximization for robust multi-robot map merging.
\newblock In \emph{2018 IEEE international conference on robotics and automation (ICRA)}, pages 2916--2923. IEEE, 2018.

\bibitem[Medioni and Kang(2004)]{medioni2004emerging}
Gerard Medioni and Sing~Bing Kang.
\newblock \emph{Emerging topics in computer vision}.
\newblock Prentice Hall PTR, 2004.

\bibitem[Murali et~al.(2022)Murali, Dutta, Gentner, Burdet, Dahiya, and Kaboli]{murali2022active}
Prajval~Kumar Murali, Anirvan Dutta, Michael Gentner, Etienne Burdet, Ravinder Dahiya, and Mohsen Kaboli.
\newblock Active visuo-tactile interactive robotic perception for accurate object pose estimation in dense clutter.
\newblock \emph{IEEE Robotics and Automation Letters}, 7\penalty0 (2):\penalty0 4686--4693, 2022.

\bibitem[Olsson et~al.(2008)Olsson, Kahl, and Oskarsson]{olsson2008branch}
Carl Olsson, Fredrik Kahl, and Magnus Oskarsson.
\newblock Branch-and-bound methods for euclidean registration problems.
\newblock \emph{IEEE Transactions on Pattern Analysis and Machine Intelligence}, 31\penalty0 (5):\penalty0 783--794, 2008.

\bibitem[Pais et~al.(2020)Pais, Ramalingam, Govindu, Nascimento, Chellappa, and Miraldo]{pais20203dregnet}
G~Dias Pais, Srikumar Ramalingam, Venu~Madhav Govindu, Jacinto~C Nascimento, Rama Chellappa, and Pedro Miraldo.
\newblock 3dregnet: A deep neural network for 3d point registration.
\newblock In \emph{Proceedings of the IEEE/CVF conference on computer vision and pattern recognition}, pages 7193--7203, 2020.

\bibitem[Pan and Chen(1999)]{pan1999complexity}
Victor~Y Pan and Zhao~Q Chen.
\newblock The complexity of the matrix eigenproblem.
\newblock In \emph{Proceedings of the thirty-first annual ACM symposium on Theory of computing}, pages 507--516, 1999.

\bibitem[Parra et~al.(2019)Parra, Chin, Neumann, Friedrich, and Katzmann]{parra2019practical}
Alvaro Parra, Tat-Jun Chin, Frank Neumann, Tobias Friedrich, and Maximilian Katzmann.
\newblock A practical maximum clique algorithm for matching with pairwise constraints.
\newblock \emph{arXiv preprint arXiv:1902.01534}, 2019.

\bibitem[Pomerleau et~al.(2013)Pomerleau, Colas, Siegwart, and Magnenat]{pomerleau2013comparing}
Fran{\c{c}}ois Pomerleau, Francis Colas, Roland Siegwart, and St{\'e}phane Magnenat.
\newblock Comparing icp variants on real-world data sets: Open-source library and experimental protocol.
\newblock \emph{Autonomous robots}, 34:\penalty0 133--148, 2013.

\bibitem[Qi et~al.(2017)Qi, Su, Mo, and Guibas]{qi2017pointnet}
Charles~R Qi, Hao Su, Kaichun Mo, and Leonidas~J Guibas.
\newblock Pointnet: Deep learning on point sets for 3d classification and segmentation.
\newblock In \emph{Proceedings of the IEEE conference on computer vision and pattern recognition}, pages 652--660, 2017.

\bibitem[Qin et~al.(2022)Qin, Yu, Wang, Guo, Peng, and Xu]{qin2022geometric}
Zheng Qin, Hao Yu, Changjian Wang, Yulan Guo, Yuxing Peng, and Kai Xu.
\newblock Geometric transformer for fast and robust point cloud registration.
\newblock In \emph{Proceedings of the IEEE/CVF conference on computer vision and pattern recognition}, pages 11143--11152, 2022.

\bibitem[Quan and Yang(2020)]{quan2020compatibility}
Siwen Quan and Jiaqi Yang.
\newblock Compatibility-guided sampling consensus for 3-d point cloud registration.
\newblock \emph{IEEE Transactions on Geoscience and Remote Sensing}, 58\penalty0 (10):\penalty0 7380--7392, 2020.

\bibitem[Rusu et~al.(2009)Rusu, Blodow, and Beetz]{rusu2009fast}
Radu~Bogdan Rusu, Nico Blodow, and Michael Beetz.
\newblock Fast point feature histograms (fpfh) for 3d registration.
\newblock In \emph{2009 IEEE international conference on robotics and automation}, pages 3212--3217. IEEE, 2009.

\bibitem[Sandryhaila and Moura(2014{\natexlab{a}})]{sandryhaila2014big}
Aliaksei Sandryhaila and Jose~MF Moura.
\newblock Big data analysis with signal processing on graphs: Representation and processing of massive data sets with irregular structure.
\newblock \emph{IEEE signal processing magazine}, 31\penalty0 (5):\penalty0 80--90, 2014{\natexlab{a}}.

\bibitem[Sandryhaila and Moura(2014{\natexlab{b}})]{sandryhaila2014discrete}
Aliaksei Sandryhaila and Jose~MF Moura.
\newblock Discrete signal processing on graphs: Frequency analysis.
\newblock \emph{IEEE Transactions on Signal Processing}, 62\penalty0 (12):\penalty0 3042--3054, 2014{\natexlab{b}}.

\bibitem[Segal et~al.(2009)Segal, Haehnel, and Thrun]{segal2009generalized}
Aleksandr Segal, Dirk Haehnel, and Sebastian Thrun.
\newblock Generalized-icp.
\newblock In \emph{Robotics: science and systems}, page 435. Seattle, WA, 2009.

\bibitem[Shi et~al.(2021)Shi, Yang, and Carlone]{shi2021robin}
Jingnan Shi, Heng Yang, and Luca Carlone.
\newblock Robin: a graph-theoretic approach to reject outliers in robust estimation using invariants.
\newblock In \emph{2021 IEEE International Conference on Robotics and Automation (ICRA)}, pages 13820--13827. IEEE, 2021.

\bibitem[Shotton et~al.(2013)Shotton, Glocker, Zach, Izadi, Criminisi, and Fitzgibbon]{shotton2013scene}
Jamie Shotton, Ben Glocker, Christopher Zach, Shahram Izadi, Antonio Criminisi, and Andrew Fitzgibbon.
\newblock Scene coordinate regression forests for camera relocalization in rgb-d images.
\newblock In \emph{Proceedings of the IEEE conference on computer vision and pattern recognition}, pages 2930--2937, 2013.

\bibitem[Speciale et~al.(2017)Speciale, Pani~Paudel, Oswald, Kroeger, Van~Gool, and Pollefeys]{speciale2017consensus}
Pablo Speciale, Danda Pani~Paudel, Martin~R Oswald, Till Kroeger, Luc Van~Gool, and Marc Pollefeys.
\newblock Consensus maximization with linear matrix inequality constraints.
\newblock In \emph{Proceedings of the IEEE Conference on Computer Vision and Pattern Recognition}, pages 4941--4949, 2017.

\bibitem[Sun(2021)]{sun2021ransic}
Lei Sun.
\newblock Ransic: Fast and highly robust estimation for rotation search and point cloud registration using invariant compatibility.
\newblock \emph{IEEE Robotics and Automation Letters}, 7\penalty0 (1):\penalty0 143--150, 2021.

\bibitem[Teed and Deng(2021)]{teed2021raft}
Zachary Teed and Jia Deng.
\newblock Raft-3d: Scene flow using rigid-motion embeddings.
\newblock In \emph{Proceedings of the IEEE/CVF conference on computer vision and pattern recognition}, pages 8375--8384, 2021.

\bibitem[Tian et~al.(2022)Tian, Luo, Zhao, and Zhou]{tian2022vibus}
Beiwen Tian, Liyi Luo, Hao Zhao, and Guyue Zhou.
\newblock Vibus: Data-efficient 3d scene parsing with viewpoint bottleneck and uncertainty-spectrum modeling.
\newblock \emph{ISPRS Journal of Photogrammetry and Remote Sensing}, 194:\penalty0 302--318, 2022.

\bibitem[Tordoff and Murray(2002)]{tordoff2002guided}
Ben Tordoff and David~W Murray.
\newblock Guided sampling and consensus for motion estimation.
\newblock In \emph{Computer Vision—ECCV 2002: 7th European Conference on Computer Vision Copenhagen, Denmark, May 28--31, 2002 Proceedings, Part I 7}, pages 82--96. Springer, 2002.

\bibitem[Torr(2002)]{torr2002bayesian}
Philip H.~S. Torr.
\newblock Bayesian model estimation and selection for epipolar geometry and generic manifold fitting.
\newblock \emph{International Journal of Computer Vision}, 50:\penalty0 35--61, 2002.

\bibitem[Watts(1999)]{watts1999networks}
Duncan~J Watts.
\newblock Networks, dynamics, and the small-world phenomenon.
\newblock \emph{American Journal of sociology}, 105\penalty0 (2):\penalty0 493--527, 1999.

\bibitem[Wu et~al.(2022)Wu, Zhao, Li, Cao, and Zha]{wu2022sc}
Xin Wu, Hao Zhao, Shunkai Li, Yingdian Cao, and Hongbin Zha.
\newblock Sc-wls: Towards interpretable feed-forward camera re-localization.
\newblock In \emph{European Conference on Computer Vision}, pages 585--601. Springer, 2022.

\bibitem[Yang et~al.(2020{\natexlab{a}})Yang, Antonante, Tzoumas, and Carlone]{yang2020graduated}
Heng Yang, Pasquale Antonante, Vasileios Tzoumas, and Luca Carlone.
\newblock Graduated non-convexity for robust spatial perception: From non-minimal solvers to global outlier rejection.
\newblock \emph{IEEE Robotics and Automation Letters}, 5\penalty0 (2):\penalty0 1127--1134, 2020{\natexlab{a}}.

\bibitem[Yang et~al.(2020{\natexlab{b}})Yang, Shi, and Carlone]{yang2020teaser}
Heng Yang, Jingnan Shi, and Luca Carlone.
\newblock Teaser: Fast and certifiable point cloud registration.
\newblock \emph{IEEE Transactions on Robotics}, 37\penalty0 (2):\penalty0 314--333, 2020{\natexlab{b}}.

\bibitem[Yang et~al.(2015)Yang, Li, Campbell, and Jia]{yang2015go}
Jiaolong Yang, Hongdong Li, Dylan Campbell, and Yunde Jia.
\newblock Go-icp: A globally optimal solution to 3d icp point-set registration.
\newblock \emph{IEEE transactions on pattern analysis and machine intelligence}, 38\penalty0 (11):\penalty0 2241--2254, 2015.

\bibitem[Yang et~al.(2021)Yang, Huang, Quan, Qi, and Zhang]{yang2021sac}
Jiaqi Yang, Zhiqiang Huang, Siwen Quan, Zhaoshuai Qi, and Yanning Zhang.
\newblock Sac-cot: Sample consensus by sampling compatibility triangles in graphs for 3-d point cloud registration.
\newblock \emph{IEEE Transactions on Geoscience and Remote Sensing}, 60:\penalty0 1--15, 2021.

\bibitem[Yu et~al.(2021)Yu, Li, Saleh, Busam, and Ilic]{yu2021cofinet}
Hao Yu, Fu Li, Mahdi Saleh, Benjamin Busam, and Slobodan Ilic.
\newblock Cofinet: Reliable coarse-to-fine correspondences for robust pointcloud registration.
\newblock \emph{Advances in Neural Information Processing Systems}, 34:\penalty0 23872--23884, 2021.

\bibitem[Zeng et~al.(2017)Zeng, Song, Nie{\ss}ner, Fisher, Xiao, and Funkhouser]{zeng20173dmatch}
Andy Zeng, Shuran Song, Matthias Nie{\ss}ner, Matthew Fisher, Jianxiong Xiao, and Thomas Funkhouser.
\newblock 3dmatch: Learning local geometric descriptors from rgb-d reconstructions.
\newblock In \emph{Proceedings of the IEEE conference on computer vision and pattern recognition}, pages 1802--1811, 2017.

\bibitem[Zhang et~al.(2023)Zhang, Yang, Zhang, and Zhang]{zhang20233d}
Xiyu Zhang, Jiaqi Yang, Shikun Zhang, and Yanning Zhang.
\newblock 3d registration with maximal cliques.
\newblock In \emph{Proceedings of the IEEE/CVF Conference on Computer Vision and Pattern Recognition}, pages 17745--17754, 2023.

\bibitem[Zhang(1994)]{zhang1994iterative}
Zhengyou Zhang.
\newblock Iterative point matching for registration of free-form curves and surfaces.
\newblock \emph{International journal of computer vision}, 13\penalty0 (2):\penalty0 119--152, 1994.

\bibitem[Zheng et~al.(2011)Zheng, Sugimoto, and Okutomi]{zheng2011deterministically}
Yinqiang Zheng, Shigeki Sugimoto, and Masatoshi Okutomi.
\newblock Deterministically maximizing feasible subsystem for robust model fitting with unit norm constraint.
\newblock In \emph{CVPR 2011}, pages 1825--1832. IEEE, 2011.

\bibitem[Zhong et~al.(2022)Zhong, You, Chen, Zhao, Sun, Zhou, Mu, Gan, and Huang]{zhong2022snake}
Chengliang Zhong, Peixing You, Xiaoxue Chen, Hao Zhao, Fuchun Sun, Guyue Zhou, Xiaodong Mu, Chuang Gan, and Wenbing Huang.
\newblock Snake: Shape-aware neural 3d keypoint field.
\newblock \emph{Advances in Neural Information Processing Systems}, 35:\penalty0 7052--7064, 2022.

\bibitem[Zhong et~al.(2023)Zhong, Zheng, Zheng, Zhao, Yi, Mu, Wang, Li, Zhou, Yang, et~al.]{zhong20233d}
Chengliang Zhong, Yuhang Zheng, Yupeng Zheng, Hao Zhao, Li Yi, Xiaodong Mu, Ling Wang, Pengfei Li, Guyue Zhou, Chao Yang, et~al.
\newblock 3d implicit transporter for temporally consistent keypoint discovery.
\newblock In \emph{Proceedings of the IEEE/CVF International Conference on Computer Vision}, pages 3869--3880, 2023.

\bibitem[Zhou et~al.(2016)Zhou, Park, and Koltun]{zhou2016fast}
Qian-Yi Zhou, Jaesik Park, and Vladlen Koltun.
\newblock Fast global registration.
\newblock In \emph{Computer Vision--ECCV 2016: 14th European Conference, Amsterdam, The Netherlands, October 11-14, 2016, Proceedings, Part II 14}, pages 766--782. Springer, 2016.

\end{thebibliography}
}

\clearpage \appendix 
\section{Appendix Section}\label{append}
\subsection{Graph Signal Processing theory}\label{theory}

\paragraph{Graph Shift} A graph $G$ can be represented in the form of $(\mathcal{V},\mathcal{A})$, where \(\mathcal{V}\) is the set of nodes \(\{v_0,v_1,\dots, v_{N-1}\}, N=|\mathcal{V}|\) and \(\mathcal{A}\in \mathbb{C}^{N\times N}\) is the graph shift, or the weighted adjacency matrix. A graph shift can reflect the connections of the graph for the edge weight is a quantitative representation between nodes. When a graph shift acts on a graph signal, it can represent the diffusion of the graph signal. A graph shift is usually normalized for proper scaling, ensuring \(||\mathcal{A}||_{spec}=1\).
\paragraph{Graph Signal}
Given a graph $G=(\mathcal{V},\mathcal{A})$, a graph signal on this graph can be seen as a map assigning each node \(v_i\) with a value \(x_i\in \mathbb{C}\). If the order of the nodes is fixed, then the graph signal is defined as a \(N\) dimensional vector \(x=(x_1,x_2,\dots,x_N)\). 
\paragraph{Graph Fourier Transform}In general, a fourier transform is the expansion of a signal on a set of bases. When performing graph fourier tansform, the signal refers to the graph signal and the bases are the eigenbasis of the graph shift, or the Jordan eigenbasis if a complete eigenbasis is not available. 
To be specific, a graph shift \(\mathcal{A}\) is eig-decomposed into $\mathcal{A}=V\Lambda V^{-1}$. The eigenvalues represent the frequencies on the graph. The fourier transform of a graph signal $x$, therefore, is defined as $\hat{x}=V^{-1}x$ and the inverse transform is $x=V\hat{x}$. $\hat{x}$ describes the content of different frequency components in the graph signal $x$.
\paragraph{Graph Filtering}
A graph filter is a type of system that accepts a graph signal as input and then generates another graph signal as output. If the graph signal is described as \(x\in \mathbb{C}^n\), then any matrix $A\in \mathbb{C}^{n\times n}$ can be seen as a graph filter and $Ax \in \mathbb{C}^n$ is the output signal. For instance, a graph shift can be a graph filter, replacing the signal value at a node with a weighted linear combination of values at its neighbors. In fact, every linear, shift-invariant graph filter can be formulated as a polynomial in the graph shift $$\mathcal{H}=\sum_{l=0}^{l=L-1}h_l\mathcal{A}_l$$ where $\mathcal{H}$ is the graph filter represented with the graph shift $\mathcal{A}$. $h_l$ is the coefficient and $L$ is the length of the graph filter.

\subsection{Problem Formulation}\label{formulation}
For two input point clouds, \(P^s\) and \(P^t\), a correspondence set \(C\) is created using either hand-crafted or learned descriptors. Each correspondence, represented by \(c\in C\), consists of a pair of points \((p^s, p^t)\), where \(p^s\) and \(p^t\) are points in \(P^s\) and \(P^t\) respectively. These correspondences can be modelled as a graph, \(G\), where each node denotes a correspondence, and the edge weights measure the compatibility between nodes. Our approach aims to decrease the size of the correspondence graph $G$ and employ the sampled correspondences for computing the 6-DoF pose transformation between \(P^s\) and \(P^t\), instead of utilising the original correspondence set $C$.

\subsection{Correspondence Generation}
Though our focus is not primarily on the creation of correspondences, it is crucial to understand how they are generated due to our approach's reliance on them. As such, we include this section.  Point cloud descriptors aim to characterise local geometry. Hand-crafted designs\cite{rusu2009fast} or deep learning methods\cite{choy2019fully}\cite{ao2021spinnet}\cite{yu2021cofinet}\cite{huang2021predator} have been employed in previous studies to create descriptors. As long as point-wise descriptors are defined, the matchability score can be calculated to form correspondences.

We denote the descriptor for point $x_i$ as $f_i$. Then the matchability score is defined using $L_2$ euclidean distance:

$$d(x_i,x_j)=||f_i-f_j||_2$$

Then for each point $x_i$, find the corresponding point $x_{k_i}$ with the highest matchability score, namely the nearest neighbor. Now a correspondence is generated:

$$c_i=(x_i,x_{k_i})$$

\subsection{Graph Construction}\label{graphconstruct}
As show in Fig.~\ref{fig:main-fig}, given a set of input correspondences $C$, we first construct a compatibility graph. Each node of the graph is a correspondence denoted as
$c_i=(x_i,y_i,z_i,u_i,v_i,w_i)$.
We further denote $p_i^s=(x_i,y_i,z_i)$ and $p_i^t=(u_i,y_i,z_i)$. As can be seen, $c_i$ is tuple of six elements, composed of coordinates of the source point $p_i^s$ and the target point $p_i^t$. We first define the \textit{distance} between correspondences as
\(
    S_{dist}(c_i,c_j)=\bigg|||p_i^s-p_j^s||-||p_i^t-p_j^t||\bigg|
\).
With this, we compute the pair-wise compatibility between correspondences, or the edge weight in the correspondence graph.
\begin{equation}
W_{ij}=\left\{
\begin{array}{rcl}
1-\frac{S_{dist}(c_i,c_j)^2}{2\times d^2_{cmp}}     & {1-\frac{S_{dist}(c_i,c_j)^2}{2\times d^2_{cmp}} > t} \\
 0    & \text{otherwise}
\end{array}\right. 
\end{equation}
where $d_{cmp}$ and $t$ are hyperparameters. We then define the adjacency matrix $W_{SOG}$ of the graph as:
\(
W_{SOG}=W\odot(W\times W)
\).In this way, we build a graph on the given correspondence set $C$. And we denote this graph as $G_{corr}$.

\subsection{Optimality Proof}\label{proof}
We claim our stochastic method is an approximation to an optimal sample operator. To prove this, we first define the object function 
\begin{equation}
    \min_{\pi}\mathbb{E}_{\Psi\sim\pi}
    ||S\Psi^T\Psi f(X)-f(X)||_2^2
\end{equation}
where $\Psi$ is the sample operator relying on $\pi$ and $S\Psi^T$ is the interpolation recovery operator. $f$ is the LapLacian high-pass filter we use and $X$ is our generalized degree signal. The optimization problem is then formulated as
\begin{equation}
\begin{aligned}
    \min_{\pi}&\mathbb{E}_{\Psi\sim\pi}
    ||S\Psi^T\Psi f(X)-f(X)||_2^2 \\
    \rm{s.t.} &\sum \pi_i=1, \pi>0
\end{aligned}
\end{equation}

To solve this, we use a Lagrange function
\begin{equation}
\begin{aligned}
    &L(\pi_i,\lambda,\mu)\\
    =& \mathbb{E}_{\Psi\sim\pi}
    ||S\Psi^T\Psi f(X)-f(X)||_2^2 \\&+\lambda(\sum \pi_i-1)+\sum \mu_i\pi_i\\
    =& ||\mathbb{E}_{\Psi\sim\pi}(S\Psi^T\Psi f(X))-f(X)||_2^2+\\
    & \mathbb{E}_{\Psi\sim\pi}||S\Psi^T\Psi f(X)-\mathbb{E}_{\Psi\sim\pi}(S\Psi^T\Psi f(X))||_2^2+\\
    &+\lambda(\sum \pi_i-1)+\sum \mu_i\pi_i\\
\end{aligned}
\end{equation}

The first item is zero, proved simply by 
\begin{equation}
\begin{aligned}
    &\mathbb{E}_{\Psi\sim\pi}(S\Psi^T\Psi f(X))_i\\
    =&\mathbb{E}_{\mathcal{M}}(\sum_{\mathcal{M}_j\in\mathcal{M}}S_{\mathcal{M}_j\mathcal{M}_j}f_{\mathcal{M}_j}(X)\delta_{\mathcal{M}_j,i})\\
    =&M\sum_k\frac{1}{M\pi_k}f_k(X)\pi_k\delta_{k,i}\\
    =&f_i(X)
    \end{aligned}
\end{equation}
where $\mathcal{M}$ is the sample set with $M$ to be its size.

The second item can be deduced to
\begin{equation}
    \begin{aligned}
        &\mathbb{E}_{\Psi\sim\pi}||(S\Psi^T\Psi f(X))_i-\mathbb{E}_{\Psi\sim\pi}(S\Psi^T\Psi f(X))_i||_2^2
    \\=&\mathbb{E}_{\mathcal{M}}(\sum_{\mathcal{M}_j, \mathcal{M}_{j'}\in\mathcal{M}}S_{\mathcal{M}_j\mathcal{M}_j}S_{\mathcal{M}_{j'}\mathcal{M}_{j'}}f_{\mathcal{M}_j}(X)^Tf_{\mathcal{M}_{j'}}(X)\\ &\delta_{\mathcal{M}_j,i}\delta_{\mathcal{M}_{j'},i})\\
    =&M^2\sum_k\frac{f_k(X)^Tf_k(X)}{M^2\pi_k^2}\pi_k\delta_{k,i}-f_i(X)^Tf_i(X)\\
    =&(\frac{1}{\pi_i}-1)f_i(X)^Tf_i(X)
    \end{aligned}
\end{equation}

Therefore, the Lagrange function can be written as 
\begin{equation}
    \sum_i(\frac{1}{\pi_i}-1)||f_i(X)||_2^2++\lambda(\sum \pi_i-1)+\sum \mu_i\pi_i
\end{equation}

By setting its derivative to zero and with the complementary slackness, we have the final result.
\begin{equation}
    \mu_i\pi_i=0 , 
    \pi_i=\frac{||f_i(X)||_2}{\sqrt{\lambda+\mu_i}}
\end{equation}

\subsection{Experimental Setup}\label{expsetup}
\paragraph{Datasets.} We consider three main datasets, i.e, the outdoor dataset KITTI\cite{geiger2012we}, the indoor dataset 3DMatch\cite{zeng20173dmatch} and its low-overlap version 3DLoMatch\cite{huang2021predator}. For KITTI, we follow the preprocess schedule of previous work\cite{bai2021pointdsc}
\cite{chen2022sc2}\cite{zhang20233d} and obtain a test set of 555 pairs of point clouds. 3DMatch is a scene-scale indoor dataset and 3DLoMatch is its subset with overlap rate ranges from 10\% to 30\%, bringing greater challenges for accurate registration.
\paragraph{Evaluation Criteria.} We follow the common evaluation criteria in 3D registration, i.e, the rotation error (RE), the translation error (TE) and the recall or success rate (RR). RE measures the angular difference between the estimated rotation matrix and the ground truth or reference rotation matrix. TE is computed as the euclidean distance between the estimated translation vector and the ground-truth, and is given in centimetres. By referring to the settings in \cite{choy2020deep}, registration is considered successful when RE$\leq15^{\circ}$ and TE$\leq30$cm on 3DMatch and 3DLoMatch datasets, and RE$\leq5^{\circ}$ and TE$\leq60$cm on the KITTI dataset. RR is then defined as the success ratio of all point cloud pairs.
\paragraph{Implementation Details.}Our FastMAC consists of a sampling process implemented in PyTorch for cuda computation and then the original Maximal Clique Registration\cite{zhang20233d} process based on C++. Our method accepts initial correspondences as input, which are generated using Fast Point Features Histograms (FPFH) \cite{rusu2009fast} and Fully Convolutional Geometric Features (FCGF) \cite{choy2019fully} as basic descriptors for both KITTI and 3DMatch\&3DloMatch. Hyperparameters like  $d_{cmp}$ and $t$ mentioned in Sec.~\ref{methods}
are set to default values, i.e, 0.1 and 0.999 respectively. The Maximal Clique Registration process remains exactly the same implemented in \cite{zhang20233d}, with the same parameter value settings. All experiments were implemented with an Intel i5-13600KF CPU and a single NVIDIA RTX4070ti. When comparing with baseline methods, we use their default parameters in their released code to ensure fairness.
\subsection{Additional Experiments}\label{addexp}
The following table demonstrates the extension result of our Time-performance Trade-off experiments. Results for RE and TE are further shown. FastMAC performs better than all previous methods when sample ratio is 50\%. With lower sample rate, RR and TE still remain competitive, with time loss decreasing dramatically, indicating its potential for real-time application.
\begin{table}[t]
\centering
\resizebox{\columnwidth}{!}{%
\begin{tabular}{cccccl}
Methods                  & ratio              & RR(\%)  & RE ($^\circ$) & TE(cm)        & Time(ms)         \\ \hline
RANSAC 1K\cite{fischler1981random}                & \multirow{9}{*}{100} & 56.58& 1.74          & 33.12         & 19.9          \\
RANSAC 10K\cite{fischler1981random}               &                      & 88.47&1.15          & 25.33         & 158.4          \\
RANSAC 100K\cite{fischler1981random}              &                      & 94.77&0.77          & 17.83         & 1549.7          \\
DGR\cite{choy2020deep}                      &                      &95.14& 0.43          & 23.28         & 330.1          \\
PointDSC\cite{bai2021pointdsc}                 &                      &96.40& 0.61          & 13.42         & 131.0          \\
PointDSC(50k)\cite{bai2021pointdsc}            &                      &96.40& 0.51          & 11.53         & 722.4          \\
PointDSC(icp)\cite{bai2021pointdsc}            &                      &96.76& 0.51          & 11.20         & 130.5          \\
SC2-PCR\cite{chen2022sc2}                  &                      &97.12& 0.41          & 9.71          & 851.1          \\
MAC\cite{zhang20233d}                      &                      &97.25& 0.36          & 8.00          & 573.0          \\ \hline
\multirow{5}{*}{FastMAC} & 50                   &\textbf{97.84}& \textbf{0.36} & \textbf{7.98} & \textbf{114.4} \\
                         & 20                   & \textbf{97.48}&\textbf{0.38} & 8.20          & \textbf{28.1} \\
                         & 10                   &\textbf{97.30}& 0.43          & 8.70          & \textbf{18.2} \\
                         & 5                    &97.12& 0.52          & 9.92          & \textbf{16.1} \\
                         & 1                    & 71.56&1.02          & 15.24         & \textbf{15.5}
\end{tabular}%
}

\caption{Comparison on RR, RE and TE with SOTA methods on the KITTI dataset with FCGF descriptor. The best results are marked in bold. $RANSAC-k$ denotes RANSAC with k iterations. PointDSC(50K): PointDSC with 50k iterations of RANSAC. PointDSC(icp): with icp refinement. }

\end{table}

\end{document}